# Benchmarking in Manipulation Research: The YCB Object and Model Set and Benchmarking Protocols

Berk Calli, *Member, IEEE,* Aaron Walsman, *Student Member,* Arjun Singh, *Student Member, IEEE, IEEE,* Siddhartha Srinivasa, *Senior Member, IEEE,* Pieter Abbeel, *Senior Member, IEEE,* and Aaron M. Dollar, *Senior Member, IEEE*

***Abstract*—In this paper we present the Yale-CMU-Berkeley (YCB) Object and Model set, intended to be used to facilitate benchmarking in robotic manipulation, prosthetic design and rehabilitation research. The objects in the set are designed to cover a wide range of aspects of the manipulation problem; it includes objects of daily life with different shapes, sizes, textures, weight and rigidity, as well as some widely used manipulation tests. The associated database provides high-resolution RGBD scans, physical properties, and geometric models of the objects for easy incorporation into manipulation and planning software platforms. In addition to describing the objects and models in the set along with how they were chosen and derived, we provide a framework and a number of example task protocols, laying out how the set can be used to quantitatively evaluate a range of manipulation approaches including planning, learning, mechanical design, control, and many others. A comprehensive literature survey on existing benchmarks and object datasets is also presented and their scope and limitations are discussed. The set will be freely distributed to research groups worldwide at a series of tutorials at robotics conferences, and will be otherwise available at a reasonable purchase cost. It is our hope that the ready availability of this set along with the ground laid in terms of protocol templates will enable the community of manipulation researchers to more easily compare approaches as well as continually evolve benchmarking tests as the field matures.**

## I. Introduction

Benchmarks are crucial for the progress of a research field, allowing performance to be quantified in order to give insight into the effectiveness of an approach. In manipulation research, particularly in robotic manipulation, prosthetic design and rehabilitation, benchmarking and performance metrics are challenging due largely to the enormous breadth of the application and task space for which researchers are working towards. The majority of research groups have therefore selected for themselves a set of objects and/or tasks that they believe are representative of the functionality that they would like to achieve/assess. Unfortunately such an approach prevents the analysis of experimental results against a common basis, and therefore makes it difficult to quantitatively interpret the performance of the described approach. Object and model sets are generally the fundamental elements involved in benchmarks for manipulation. Substantial effort has already been put into providing mesh model databases of objects (e.g. [1-4] with a thorough overview provided in section II), generally for object recognition or planning purposes. There have, however, been very few instances of proposed object/task sets for which the physical objects are available to researchers. Access to the objects is crucial to performance benchmarking as many aspects of the manipulation process cannot be modeled, thereby requiring experiments to demonstrate success or examine failure modes.

In this paper, we present an object set for manipulation research in the fields of robotics, prosthetic design and rehabilitation, a framework for standard task protocols, and a number of example protocols along with experimental implementation of them. The object set is specifically designed to allow for widespread dissemination of the physical objects and manipulation scenarios selected based on a survey of the most common objects utilized in research in the target fields along with a number of additional practical constraints. Along with the physical objects, textured mesh models and high quality images are provided together with their physical properties to enable realistic simulations. These models are integrated into the MoveIt motion planning tool [5] and the ROS manipulation stack. The set will be freely distributed to research groups worldwide at a series of tutorials at robotics conferences, and will be otherwise available at a reasonable purchase cost. Our goal is to do as much as possible to facilitate the widespread usage of a common set of objects and tasks in order to allow easy comparison of results between research groups worldwide.

In choosing the set of objects and data provided, a number of issues were considered. The objects should span a variety of shapes, sizes, weight, rigidity and texture, as well as span a wide range of manipulation applications and challenges. Still, several practical constraints must be considered, including the number and size of the objects to allow for easy shipping and storage, keeping the overall cost reasonable, providing objects that are durable so as to not substantially degrade over time or with usage, as well as to choose objects that are likely to be available in a similar form in the future. Preliminary data in the repository includes high-resolution 3D point cloud data with associated meshes and texture (visual) information, object mechanical properties such as major dimensions and mass, as well as models for integration into planning and simulation software, all available at: http://rll.eecs.berkeley.edu/ycb/.

In addition to the object and model set, we provide a systematic approach to define manipulation protocols and benchmarks using the set. Guidelines for designing well-defined manipulation protocols and benchmarks are provided through a template for easy protocol and benchmark design. Furthermore, using this template, example protocols and benchmarks are derived for the assessment of various aspects of robotic manipulation i.e. mechanical design, manipulation planning, dexterity and learning. The implementation of these

Funding for this work was provided in part by the National Science Foundation, grants IIS- 0953856, IIS-1139078, and IIS-1317976.

B. Calli and A. Dollar are with the Department of Mechanical Engineering and Materials Science, Yale University, New Haven, CT, USA (203-436-9122; e-mail: {berk.calli; aaron.dollar}@yale.edu).

A. Singh and P. Abbeel are with the Department of Computer Science, University of California at Berkeley, Berkeley, CA, USA (e-mail: arjun810@gmail.com; pabbeel@cs.berkeley.edu).

A. Walsman and S. Srinivasa are with the Robotics Institute, Carnegie Mellon University, Pittsburgh, PA, USA (e-mail: aaronwalsman@gmail.com; ss5@andrew.cmu.edu).

benchmarks on real robotic systems are also provided to demonstrate the benchmarks' abilities to quantitatively evaluate the manipulation capabilities of various systems.

We expect to continually expand this work by including additional data on the objects (i.e. inertial properties), by proposing additional benchmarks for manipulation, but more importantly, by creating a web portal for the user community to engage in this effort, proposing changes to the object set and putting forth their own protocols and benchmarks, among others.

The remainder of this paper is organized as follows: First a comprehensive literature survey on object sets and benchmarks is presented in Section II. Following that our object set is presented and explained in Section III. In Section IV, guidelines are provided for designing protocols and benchmarks. In Section V, the example protocols and benchmarks are presented. The paper is concluded with discussions and future work in Section VI.

## II. Related Work

For benchmarking in manipulation, specifying an object set is essential. Table 1 summarizes the object sets proposed in prior work. Even though there are a large number of efforts that provide datasets of object mesh models which are useful for many simulation and planning applications, as well as for benchmarking in shape retrieval, these datasets have limited utility for manipulation benchmarking due to several reasons: First, since most of them are not designed specifically for manipulation benchmarking, the selected objects do not usually cover the shape and function variety needed for a range of manipulation experiments. Second, none of these databases provide the objects' physical properties, which are necessary to conduct realistic simulations. Lastly and perhaps most importantly, the vast majority of objects in these sets are not easily accessible by other researchers, preventing their use in experimental work. Exceptions to this include [6], which provides an online shopping list (but it is outdated with many dead links), the recently-announced Amazon Picking Challenge [7], which provides a shopping list to purchase objects meant for a narrow bin-picking task, and commercial kits available for some rehabilitation-focused tests [8-11] that provide limited set of objects but are not representative of a wide range of manipulation tasks. The current effort is unique in that it provides a large amount of information about the objects necessary for many simulation and planning approaches, makes the actual objects readily available for researchers to utilize experimentally, and includes a wide range of objects to span many different manipulation applications.

In the following sections, we provide a detailed overview of prior related benchmarking efforts, discussing their scope and limitations. For organization purposes, we first discuss work primarily related to robotic manipulation (including vision and learning applications), then efforts in rehabilitation, including prosthetics.

### A. *Robotic Manipulation:*

The necessity of manipulation benchmarks is highly recognized in the robotics community [12-14] and continues to be an active topic of discussion at workshops on robotic manipulation (e.g. [15]). As mentioned earlier, the majority of prior work related to object sets has involved just the models and images of those objects (with varying degrees of information, from purely shape information to textural plus shape), often created for research in computer vision (e.g. [2, 16, 17]). There have also been a number of shape/texture sets designed for/by the robotics community, particularly for applications such as planning and learning. The Columbia Grasp Database (CGD) [3] rearranges the object models of the Princeton Shape Benchmark (PSB) [18] for robotic manipulation and provides mesh models of 8000 objects together with assigned successful grasps per model. Such a database is especially useful for implementing machine learning-based grasp synthesis algorithms in which large amounts of labeled data are required for training the system. A multi-purpose object set which also targets manipulation is the KIT Object Models Database [19] which provides stereo images and textured mesh models of 100 objects. While there are a large number of objects, the shape variety is limited, and like the previously mentioned datasets, the objects are not easily accessible to other researchers.

There have only been two robotics-related efforts in which the objects are made relatively available. The household objects list [6] provides good shape variety that is appropriate for manipulation benchmarking, as well as a shopping list for making the objects more easily accessible to researchers. Unfortunately, the list is outdated, and most objects are not available anymore. Also, the 3D models of the objects are not supplied which prevents the use of the object set in simulations. Very recently, the Amazon Picking Challenge [7] also provides a shopping list for items, but those were chosen specific to the bin-picking application and do not have models associated with them.

In terms of other robotic manipulation benchmarking efforts, a number of simulation tools have been presented in the literature. The OpenGRASP benchmarking suite [20] presents a simulation framework for robotic manipulation. The benchmarking suite provides test cases and setups, and a standard evaluation scheme for the simulation results. So far, a benchmark for grasping known objects has been presented using this suite. VisGraB [21] provides a benchmark framework for grasping unknown objects. The unique feature of this software is utilizing real stereo images of the target objects for grasp synthesis, and executing and evaluating the result in a simulation environment. For gripper and hand design, benchmark tests [22, 23] are proposed for evaluating the ability of the grippers to hold an object, but only cylindrical objects are used.

**Table I: Object Datasets in Literature (Sorted by Year)**

| | Dataset name | Year | Data Type | Purpose | Object number / Category | Physical objects available ? | Website |
|---|---|---|---|---|---|---|---|
| 1 | BigBIRD [1] | 2014 | Meshes with texture + HQ images | Object recognition | 100 | No | http://rll.eecs.berkeley.edu/bigbird |
| 2 | Amazon Picking Challenge [7] | 2014 | Shopping list | Grasping | 27 | Yes | http://amazonpickingchallenge.org/ |
| 3 | SHREC'14 [2] | 2014 | Mesh models | Object retrieval | 8987 / 171 | No | http://www.itl.nist.gov/iad/vug/sharp/contest/2014/Generic3D/ |
| 4 | SHREC'12 [24] | 2012 | Mesh models | Object retrieval | 1200 / 60 | No | http://www.itl.nist.gov/iad/vug/sharp/contest/2012/Generic3D/ |
| 5 | The KIT object models database [19] | 2012 | Mesh with texture, stereo images | Recognition, localization and manipulation | 100 | No | http://i61p109.ira.uka.de/ObjectModelsWebUI/ |
| 6 | VisGraB [21] | 2012 | Stereo images | Manipulation | 18 | No | http://www.robwork.dk/visgrab/ |
| 7 | The Object Segmentation Database [17] | 2012 | RGB-D images | Object segmentation | N/A | No | http://users.acin.tuwien.ac.at/arichtsfeld/?site=4 |
| 8 | Toyohashi shape benchmark [25] | 2012 | Mesh models | Object retrieval | 10k / 352 | No | http://www.kde.cs.tut.ac.jp/benchmark/tsb/ |
| 9 | The Willow Garage Object Recognition Challenge [26] | 2012 | RGB-D images | Object recognition | N/A | No | http://www.acin.tuwien.ac.at/forschung/v4r/mitarbeiterprojekte/willow/ |
| 10 | SHREC'11 [27] | 2011 | Mesh models | Object retrieval | 600 | No | http://www.itl.nist.gov/iad/vug/sharp/contest/2011/NonRigid/ |
| 11 | Berkeley 3-D Object Dataset [28] | 2011 | RGB-D dataset of room scenes | Object detection | N/A | No | http://kinectdata.com/ |
| 12 | RGB-D Object Dataset [29] | 2011 | RGB-D dataset | Object detection and recognition | 300 / 51 | No | http://rgbd-dataset.cs.washington.edu/ |
| 13 | The OpenGRASP benchmarking suite [20] | 2011 | Mesh with texture, stereo images | Grasping | Uses KIT database | No | http://opengrasp.sourceforge.net/benchmarks.html |
| 14 | SHREC'10 [30] | 2010 | Mesh models | Object retrieval | 3168 / 43 | No | http://tosca.cs.technion.ac.il/book/shrec_robustness2010.html |
| 15 | The Columbia Grasp Database [3] | 2009 | Mesh models | Grasping | ~8000 | No | http://grasping.cs.columbia.edu/ |
| 16 | Benchmark Set of Domestic Objects [6] | 2009 | Shopping list | Robotic manipulation | 43 | Yes | http://www.hsi.gatech.edu/hrl/object_list_v092008.shtml |
| 17 | Bonn Architecture Benchmark [31] | 2009 | Mesh models | Object retrieval | 2257 | No | ftp://ftp.cg.cs.uni-bonn.de/pub/outgoing/ArchitectureBenchmark |
| 18 | Engineering Shape Benchmark [32] | 2008 | Mesh models | Object retrieval | 867 | No | https://engineering.purdue.edu/PRECISE/shrec08 |
| 19 | [3D Object Retrieval Benchmark] [33] | 2008 | Mesh models | Object retrieval | 800 / 40 | No | http://www.itl.nist.gov/iad/vug/sharp/benchmark/ |
| 20 | McGill 3D Shape Benchmark [34] | 2008 | Mesh models | Shape retrieval | N/A | No | http://www.cim.mcgill.ca/~shape/benchMark/ |
| 21 | The Toronto Rehabilitation Institute Hand Function Test [35] | 2008 | Commercial Kit / No model data | Prosthetics and Rehabilitation | 14 | No | http://www.rehabmeasures.org/Lists/RehabMeasures/PrintView.aspx?ID=1044 |
| 22 | GRASSP [9] | 2007 | Commercial Kit / No model data | Prosthetics and Rehabilitation | N/A | Yes | http://grassptest.com/ |
| 23 | AIM@SHAPE Shape Repository [16] | 2006 | Mesh models | General | 1180 | No | http://shapes.aim-at-shape.net/viewmodels.php |
| 24 | The Princeton Shape Benchmark [18] | 2004 | Mesh models | Shape-based retrieval | 1,814 | No | http://shape.cs.princeton.edu/benchmark/ |
| 25 | [Mesh Deformation Dataset] [36] | 2004 | Mesh models | Mesh transformation | N/A / 13 | No | http://people.csail.mit.edu/sumner/research/deftransfer/data.html |
| 26 | NTU 3D model benchmark [37] | 2003 | Mesh models | Shape retrieval | 1,833 | No | http://3d.csie.ntu.edu.tw/ |
| 27 | SHAP [8] | 2002 | Commercial Kit / No model data | Prosthetics and Rehabilitation | - | Yes | http://www.shap.ecs.soton.ac.uk/ |
| 28 | Action Research Arm Test [10] | 1981 | Commercial Kit / No model data | Prosthetics and Rehabilitation | 19 | Yes | http://saliarehab.com/actionresearcharmtestarat.html |
| 29 | Jebsen-Taylor Hand Function Test [11] | 1969 | Commercial Kit / No model data | Prosthetics and Rehabilitation | N/A | Yes | N/A |
| 30 | The ITI database [38] | N/A | Mesh models | Object retrieval | 544 / 13 | No | http://vcl.iti.gr/3d-object-retrieval/ |
| 31 | Model Bank Library [39] | N/A | Mesh with texture | General | 1200 | No | http://digimation.com/3d-libraries/model-bank-library/ |
| 32 | SketchUp [4] | N/A | Mesh with and w/o texture | General | N/A | No | https://3dwarehouse.sketchup.com/ |
| 33 | Robocup @home [40] | Multi. | no data | Manipulation | N/A | No | http://www.robocupathome.org/ |

## B. *Prosthetics and Rehabilitation*

In the general field of rehabilitation, including prosthetics, there are a number of evaluation tools used by therapists to attempt to quantify upper-limb function. Some of these are common, commercially-available, and have been substantially published on, including "normative" data to compare the patient's performance to others. Other tests have only been proposed in the literature and not (yet, at least) widely utilized.

The tests that have a commercial setup available are Box and Blocks Test [41], 9-hole-peg test [42], Jebsen-Taylor Hand Function Test [11], Action Research Arm Test (ARAT) [10], the Southampton Hand Assessment Procedure (SHAP) [8] and Graded Redefined Assessment of Strength, Sensibility and Prehension (GRASSP) test [9]. The setups for Box and Blocks Test and 9-hole-peg tests are very specific for these protocols, including timed movements of simple objects. The setup for Jebsen-Taylor Hand Function Test also consists of quite limited set of objects that are specific to limited tasks. ARAT assesses upper limb function and its commercial set [43] contain objects such as wooden blocks of various sizes, glasses, a stone, a marble, washers and bolts. The test proposes actions like placing a washer over a bolt and pouring water from a glass to another. Again, this set is far from providing an adequate object variety for deriving manipulation benchmarks. GRASSP measure has been proposed for the assessment of upper limb impairment. This measure is based on a commercial kit available in [44]. Apart from a specialized manipulation setup, the kit also includes 9-hole peg test, jars and a bottle. The SHAP setup includes some objects of daily living such as a bowl, a drink carton, and a jar, together with some geometrical shapes, for which subjects are requested to do a variety of manipulation tasks, including pouring the drink, opening the jar etc. Despite enabling a larger possibility of manipulation tasks than the previously mentioned setups, the GRASSP and SHAP setups are still bounded to a limited number of tasks, and both are pricey.

Some well-known tests that do not provide a commercial setup are Grasp and Release Test [45], The Toronto Rehabilitation Institute Hand Function Test [35] and Activities Measure for Upper Limp Amputees (AM-ULA) [46]. The Grasp and Release Test is proposed for evaluating the performance of neuroprosthetic hands. For this test, detailed descriptions of the objects are given, but these objects are not easily obtainable, and the set includes an outdated object i.e. a videotape. The Toronto Rehabilitation Institute Hand Function Test (also known as Rehabilitation Engineering Laboratory Hand Function Test [47]) evaluates palmer (power) and lateral (precision) grasp abilities of individuals by using an object set consist of a mug, a book, a paper, a soda can, dices, a pencil etc. Even though it is claimed that the objects used in this test are easily obtainable, maintaining the exact object definition is hard and one of the objects is an outdated cellular phone. AM-ULA defines several quality measures for assessing the manipulation tasks, and various daily activities are proposed for the assessment. The objects used in these activities are not standardized.

In addition to these test, some works in literature use their own setups for assessment. In [48], tasks such as "use a hammer and nail", "stir a bowl", "fold a bath towel", "use a key in a lock" are proposed for evaluating upper limb prosthesis. In [49], the performance of the neuroprosthesis are evaluated by asking the patient to perform grasping and lifting tasks, as well as phone dialing, pouring liquid from a pitcher and using spoon and fork. In [50], for evaluating the outcomes of a protocol for stoke rehabilitation, blocks, Lego and pegs are used together with daily life activities like folding, buttoning, pouring and lifting. In [51], the outcomes of the neuroprosthesis are measured with Box and Blocks Test and Clothes Pin Relocation Task together with the evaluation of actions of daily living i.e. using a fork and a knife, opening a jar, stirring a spoon in a bowl. In none of the abovementioned assessment procedures, the descriptions of the objects are provided, however.

In our object set, we have included the objects that are commonly used in these assessment procedures (i.e. a mug, a bowl, a pitcher, washers, bolts, kitchen inventories, pens, key-padlock etc.). We also included objects that will allow designing protocols which focus on activities of daily living. Moreover, widely used manipulation tests such as 9-hole peg, box and blocks and clothes peg allocation are also provided.

# III. The Object and Data Set

The proposed object set can be seen in Figures 1-7 and listed in Table I. In this section, we describe the object set and the reasoning behind the choices (section III.A), a description of the process and data involved in the scans of the objects (III.B), and the models and integration into simulation and planning packages (III.C) and a brief functional demonstration (III.D).

## A. *Objects*

We aimed to choose objects that are frequently used in daily life, and went through the literature to take into account the objects that are frequently used in simulations and experiments. We also benefit from the studies on objects of daily living [52] and daily activities checklist such as [53].

In compiling the proposed object and task set, we needed to take a number of additional practical issues into consideration:

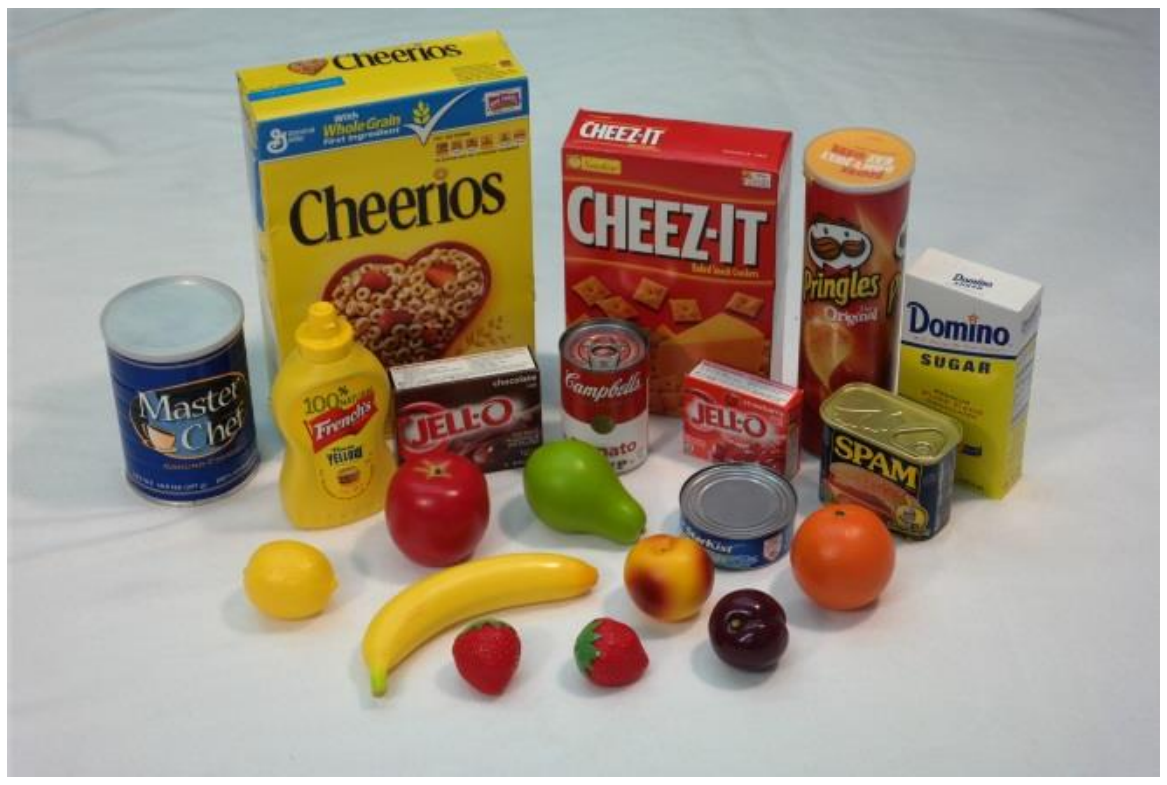

**Fig. 1: Food items in the YCB Object Set: back row: coffee can, cereal box, cracker box, chips can, box of sugar; middle row: mustard container, chocolate pudding box, tomato soup can, gelatin box, potted meat can; front: plastic fruit (lemon, apple, pear, orange, banana, peach, strawberries, plum).**

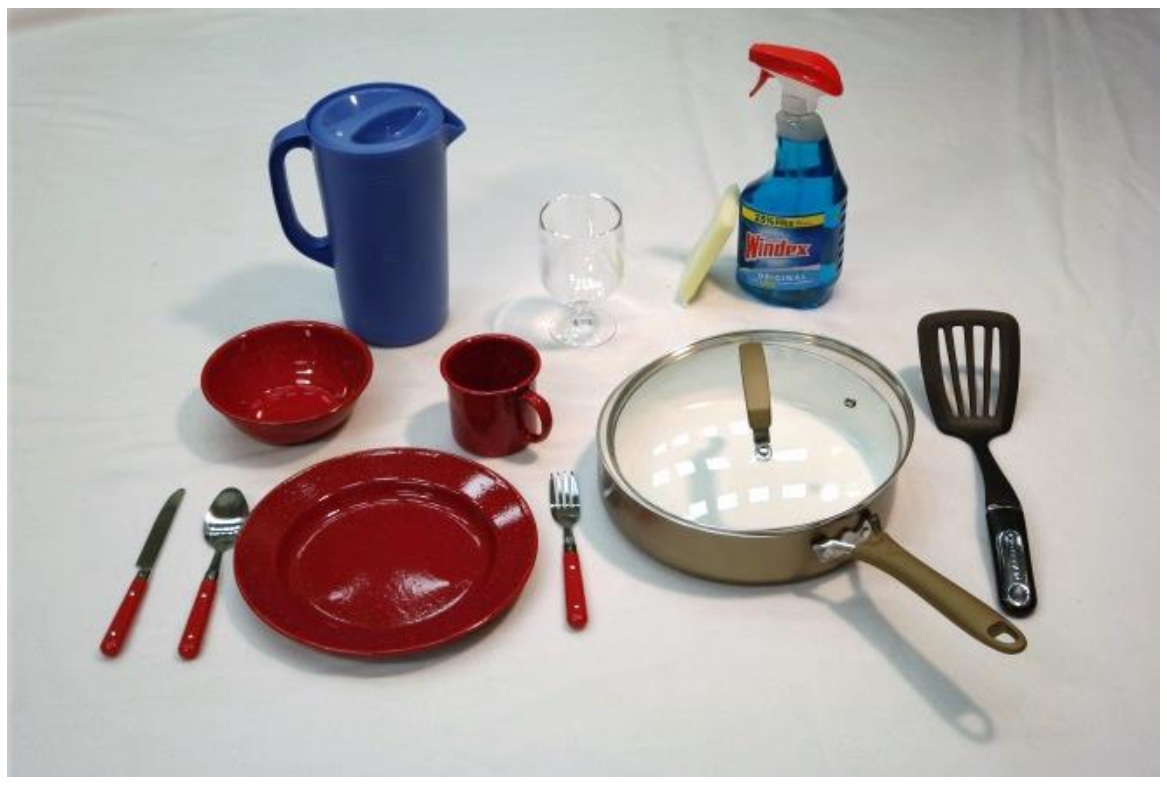

**Fig. 2: Kitchen items in the YCB Object Set: back row: pitcher, plastic wine glass, abrasive sponge, glass cleaner; front: enamel-coated metal dishware (bowl, coffee mug and plate), eating utensils (knife, spoon, and fork), cooking skillet with glass lid, plastic turner.**

• *Variety:* In order to cover as many aspects of robotic manipulation as possible, we included objects that have a wide variety of shape, size, transparency, deformability, and texture. Grasping and manipulation difficulty was also a criterion: for instance, some objects in the set are well approximated by simple geometric shapes and relatively easy for grasp synthesis and execution, while other objects have higher shape complexity and more challenging for grasp synthesis and execution.

• *Use:* We included objects that are not only interesting for grasping, but also have a range of manipulation uses. For example, a pitcher and a cup; nails and a hammer; pegs, cloths and rope. We also included "assembly" items/tasks: a set of children's stacking cups, a toy airplane (Fig. 6) that must be assembled and screwed together and Lego (Fig. 7). Additionally, widely used standard manipulation tests in rehabilitation, such as the box and blocks [41] and 9-hole-peg test [42], are included. As above, these tasks are intended to span a wide range of difficulty, from relatively easy to very difficult. Furthermore, the ability to quantify task performance was also prioritized, including aspects such as level of difficulty, time-to-completion, and success rate, among others.

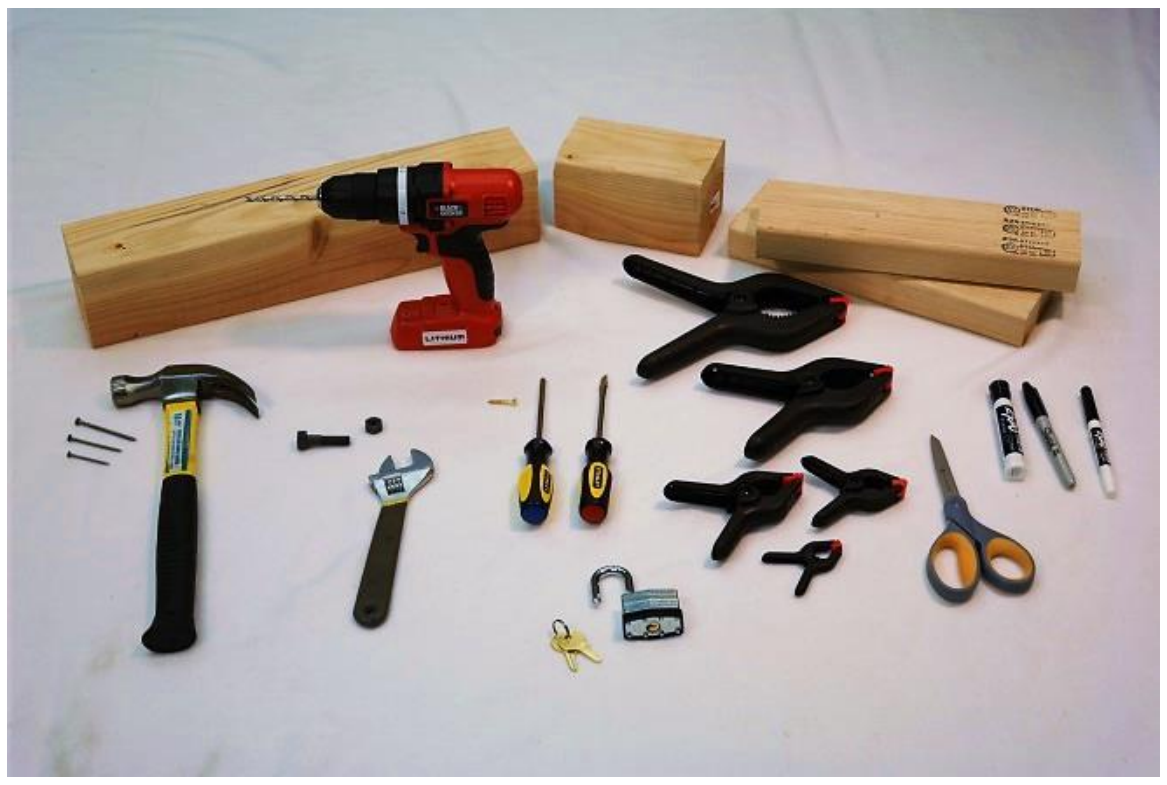

**Fig. 3: Tool items in the YCB Object Set: back: wood blocks (four items, three sizes), power drill; front: nails (three sizes), hammer, plastic bolt and nut; adjustable wrench, wood screw, phillips and flat screwdrivers, padlock and keys, spring clamps (five sizes), scissors, markers (three sizes/types).**

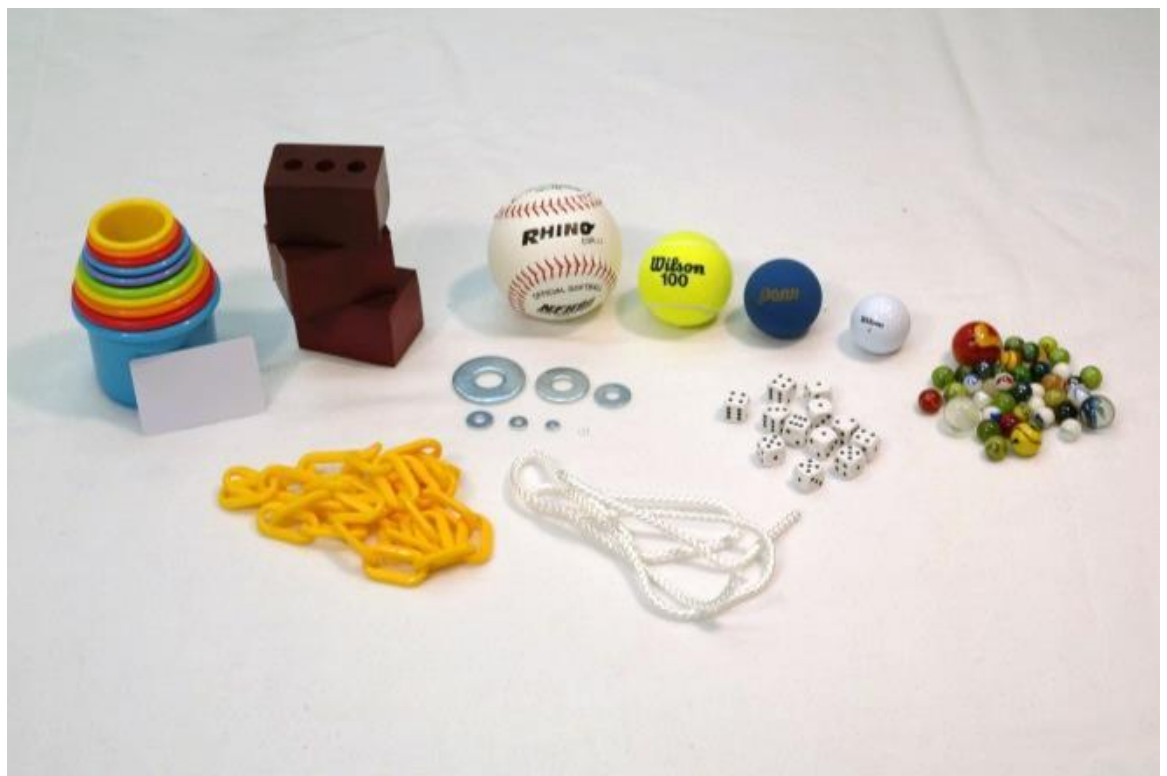

**Fig. 4: Shape items in the YCB Object Set: back: stacking blocks (set of 10), credit card blank, foam bricks (set of 3), spherical objects (softball, tennis ball, racquetball, golf ball, marble assortment); front: plastic chain, nylon rope, washer assortment (7 sizes), dice (set of 12).**

• *Durability:* We aimed for objects that can be useful long term, and therefore avoid objects that are fragile or perishable. Also, to increase the longevity of the object set, we chose the objects that are likely to remain in circulation and change relatively little in the near future.

• *Cost:* We aimed to keep the cost of the object set as low as possible to broaden accessibility. We therefore selected standard consumer products, rather than, for instance, custom-fabricated objects and tests. Current cost for the objects is approximately $350.

• *Portability:* We aimed to have an object set that fits in a large-sized suitcase and be below the normal airline weight limit (22kg) in order to allow easy shipping and storage.

After considering these practical issues and reviewing the literature, the final objects were selected (Table II, Figs. 1-7). The objects in the set can be divided into the following categories: food items, kitchen items, tool items, shape items, task items. Objects from ID 1 to 19 are the food items, containing real boxed and canned items, as well as wooden fruits of complex shapes. The objects from ID 20 to 33 are kitchen items, containing objects for food preparation and

**Table II: Object Set Items and Properties**

| ID | Class | Object | Mass | Dims. (mm) |
|---|---|---|---|---|
| 1 | Food items | Cereal Box | 510g | 80 x 195 x 295 |
| 2 | | Cracker Box | 453g | 60 x 160 x 230 |
| 3 | | Sugar Box | 514g | 38 x 89 x 175 |
| 4 | | Pudding Box | 187g | 35 x 110 x 89 |
| 5 | | Gelatin Box | 97g | 28 x 85 x 73 |
| 6 | | Potted Meat Can | 370g | 50 x 97 x 82 |
| 7 | | Master Chef Can | 414g | 102 x 139 |
| 8 | | Tuna fish can | 171g | 85 x 33 |
| 9 | | Chips Can | 205g | 75 x 250 |
| 10 | | Mustard Bottle | 431g | 50 x 85 x 175 |
| 11 | | Tomato Soup Can | 349g | 66 x 101 |
| 12 | | Banana | 66g | 36 x 190 |
| 13 | | Strawberry | 18g | 43.8 x 55 |
| 14 | | Apple | 68g | 75 |
| 15 | | Lemon | 29g | 54 x 68 |
| 16 | | Peach | 33g | 59 |
| 17 | | Pear | 49g | 66.2 x 100 |
| 18 | | Orange | 47g | 73 |
| 19 | | Plum | 25g | 52 |
| 20 | Kitchen Items | Windex Bottle | 1022g | 80 x 105 x 270 |
| 21 | | Sponge | 6.2g | 72 x 114 x 14 |
| 22 | | Pitcher Base | 178g | 108 x 235 |
| 23 | | Pitcher Lid | 66g | 123 x 48 |
| 24 | | Plate | 279g | 258 x 24 |
| 25 | | Bowl | 147g | 159 x 53 |
| 26 | | Fork | 34g | 14 x 20 x 198 |
| 27 | | Spoon | 30g | 14 x 20 x 195 |
| 28 | | Knife | 31g | 14 x 20 x 215 |
| 29 | | Slotted Turner | 105g | 25 x 35 x 410 |
| 30 | | Wine glass | 133g | 89 x 137 |
| 31 | | Mug | 118g | 80 x 82 |
| 32 | | Skillet | 950g | 270 x 25 x 30 |
| 33 | | Skillet Lid | 652g | 270 x 10 x 22 |
| 34 | Tool Items | Scissors | 82g | 87 x 200 x 14 |
| 35 | | Permanent Marker | 10g | 11.5 x 137 |
| 36 | | S Erasable Marker | 8.3g | 10.5 x 136 |
| 37 | | L Erasable Marker | 16g | 19 x 119 |
| 38 | | Keys | 6g | 23 x 43 x 2.2 |

**Table II (cont): Object Set Items and Properties**

| ID | Class | Object | Mass | Dims. (mm) |
|---|---|---|---|---|
| 39 | Tool Items | Padlock | 208g | 24 x 47 x 65 |
| 40 | | Hammer | 688g | 32 x 40 x 160 |
| 41 | | Nails | [2,2.7,4.8] g | [4x25, 3x53, 4x63] |
| 42 | | Phillips Screwdriver | 70g | 23 x 198 |
| 43 | | Flat Screwdriver | 70g | 23 x 198 |
| 44 | | Adjustable Wrench | 247g | 10 x 26 x 204 |
| 45 | | 2x4 Wood block | 414g | 38 x 90 x 300 |
| 46 | | 4x4 Wood block, long | 1720g | 90 x 90 x 410 |
| 47 | | 4x4 Wood block, short | 638 | 90 x 90 x 152 |
| 48 | | XS Clamp | 8.3g | 50 x 65 x 11 |
| 49 | | S Clamp | 30g | 80 x 91 x 25 |
| 50 | | M Clamp | 59g | 90 x 115 x 27 |
| 51 | | L Clamp | 125g | 125 x 165 x 32 |
| 52 | | XL Clamp | 202g | 165 x 213 x 37 |
| 53 | | Power Drill | 895g | 35 x 46 x 184 |
| 54 | Shape Items | Credit Card blank | 5.2g | 54 x 85 x 1 |
| 55 | | Soft Ball | 191g | 96 |
| 56 | | Tennis Ball | 58g | 64.7 |
| 57 | | Racquetball | 41g | 55.3 |
| 58 | | Golf Ball | 46g | 42.7 |
| 59 | | S Marble | 3.6g | 14 |
| 60 | | M Marble | 5.3g | 16 |
| 61 | | L Marble | 20g | 24.7 |
| 62 | | XL Marble | 59g | 35.2 |
| 63 | | Cups | [13,14,17,19, 21,26,28,31,35,38] g | [55x60, 60x62, 65x64, 70x66, 75x68, 80x70, 85x72, 90x74, 95x76, 100x78] |
| 64 | | Foam Brick | 28g | 50 x 75 x 50 |
| 65 | | Dice | 5.2g | 16.2 |
| 66 | | Washers | [0.1,0.7,1.1,3,5.3,19,48] g | [6.4, 10, 13.3, 18.8, 25.4, 37.3, 51] |
| 67 | | Rope | 20g | 6.4 x 1440 |
| 68 | | Chain | 100g | 2 x 4 x 130 |
| 69 | Task Items | Clear Box | 302g | 292 x 429 x 149 |
| 70 | | Box Lid | 159g | 292 x 429 x 20 |
| 71 | | Colored Wood Blocks | 10.8g | 26 |
| 72 | | 9-Peg-Hole Test | 1435g | 1150 x 1200 x 1200 |
| 73 | | Toy Airplane | 570g | 171 x 266 x 280 |

serving, as well as glass cleaner and a sponge. The objects from 34 to 53 form the tool category, containing not only common tools, but also items such as nails, screws, and wood to utilize them. The shape items are from ID 54 to 68, which span a range of sizes (spheres, cups, and washers), as well as compliant objects such as foam bricks, rope, and chain. The manipulation task items are the objects with IDs 69 to 73, and include two widely used tasks in rehabilitation benchmarking (box-and-blocks [41] and 9-hole peg test [42]) as well as a simple and a complex assembly task (Lego and children's airplane toy respectively). We also include a timer in the kit, which not only provides accurate timing of the task, but also can specify the initial position of the hand at the beginning of the task execution.

While there are an unlimited number of manipulation tasks that might be able to be done with these objects, we provide some examples for each category in Table III (with in-depth discussion of tasks and protocols in Section IV).

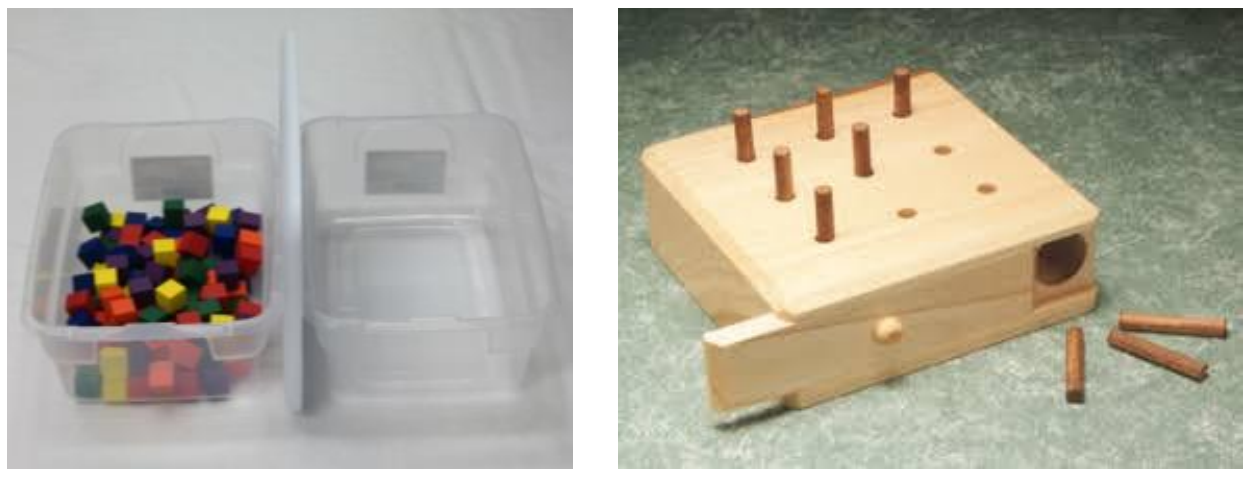

**Fig. 5: (left) Box-and-blocks test objects: set of 100 wooden cubes, two containers and height obstacle (container lid) between them. (right) 9-hole peg test: wooden pegs are placed in**

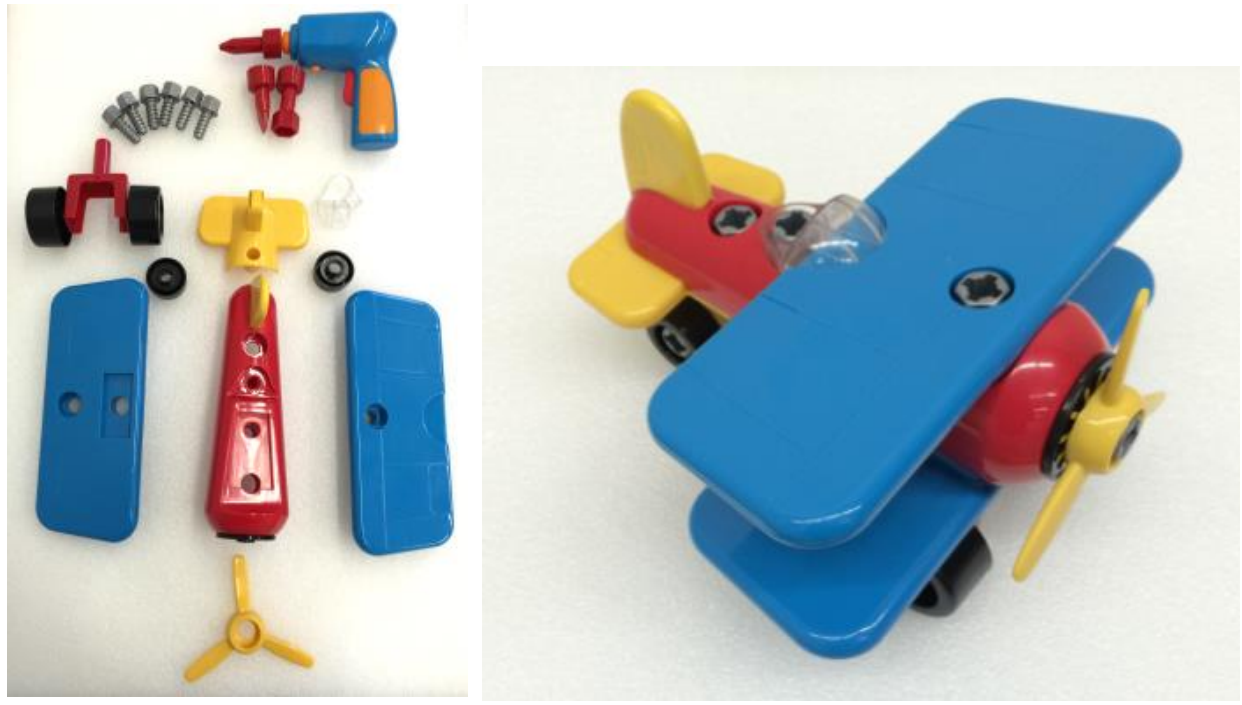

**Fig. 6: Assembly object: toy airplane disassembled (left), including toy power screwdriver, and fully assembled (right).**

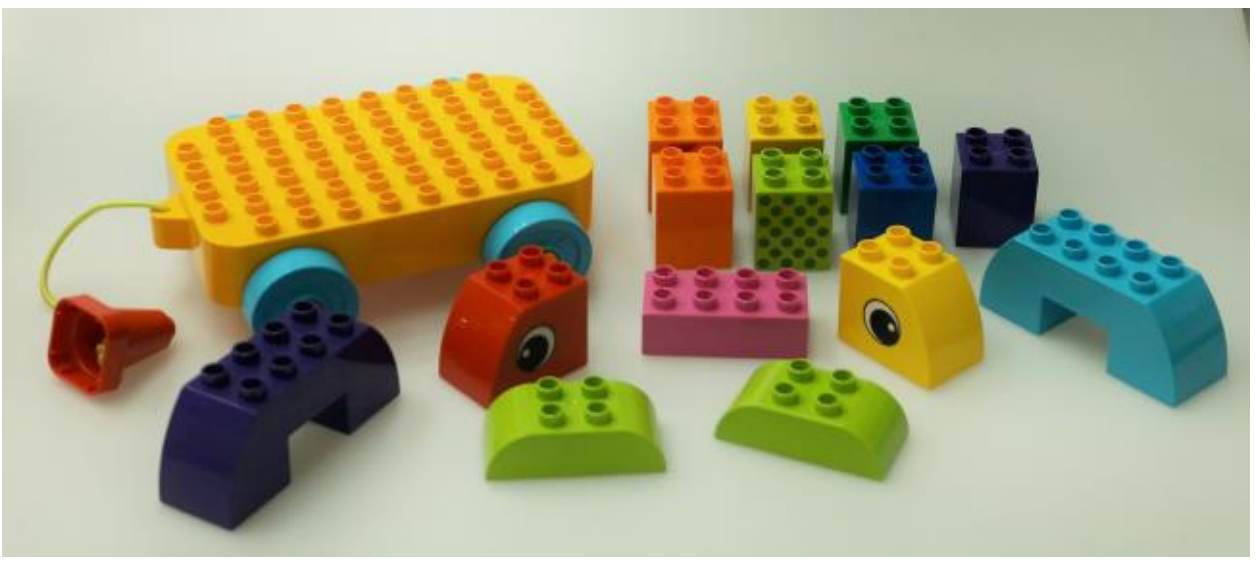

**Fig. 7: Assembly object: Lego pieces.**

### B. *Scans*

In order to ease adoption, we collect visual data that is commonly required for grasping algorithms and generate 3D models for use in simulation. We use the scanning rig used to collect the BigBIRD dataset [1]. The rig, shown in Fig. 8, has 5 RGBD sensors and 5 high-resolution RGB cameras arranged in a quarter-circular arc. Each object is placed on a computer-controlled turntable, which is rotated by 3 degrees at a time, yielding 120 turntable orientations. Together, this yields 600 RGBD images and 600 high-resolution RGB images. The process is completely automated, and the total collection time for each object is under 5 minutes.

We then use Poisson surface reconstruction to generate watertight meshes. Afterwards, we project the meshes onto each image to generate segmentation masks. Note that Poisson reconstruction fails on certain objects with missing depth data; specifically, transparent or reflective regions of objects usually do not register depth data. We will later provide better models for these objects using algorithms that take advantage of the high-resolution RGB images for building models.

**Table III Suggestions for Manipulation Tasks**

| Object Category | Suggested Tasks |
|---|---|
| Food items | • Packing/unpacking the groceries. |
| Kitchen items | • Table setting,<br>• Wipe down table with sponge and Windex,<br>• Cooking scenarios. |
| Tool items | • Nailing,<br>• Drilling,<br>• Unlocking the pad using the key,<br>• Placing the pegs on the rope.<br>• Unscrewing a bolt using the wrench,<br>• Cutting a paper with the scissors,<br>• Writing on a paper.<br>• Screwing the nut on the bolt. |
| Shape items | • Sorting marbles into the plastic blocks,<br>• Unstacking/stacking the cups,<br>• Placing the washers onto the bolt. |
| Other items | • Box and blocks test,<br>• Toy plane assembly/disassembly,<br>• 9-peg hole tests,<br>• Lego assembly/disassembly. |

In total, for each object, we provide:

- 600 RGBD images
- 600 high-resolution RGB images
- Segmentation masks for each image
- Calibration information for each image
- Texture-mapped 3D mesh models

The object scans can be found at [54].

### C. *Models*

Based on the scans of the objects, there are several ways in which object models can be easily integrated into a variety of robot simulation packages. For example, in the MoveIt [5] simulation package, the mesh can be used as a collision object directly. Furthermore, a Unified Robot Description Format (URDF) file can be automatically constructed to integrate with ROS [55]. This provides a way of specifying mass properties, and can link to alternate representations of the mesh for visualization and collision. Integration with the OpenRAVE [56] simulation package is similarly straight-forward where we link to the display and collision meshes from a KinBody XML file. Using the scans, we have created URDF and KinBody files for all of the objects in the dataset, provided alongside the scans at [54].

Once in a simulation environment, a variety of motion planners and optimizers can use these models either as collision or manipulation objects. Some algorithms, such as CHOMP [57], require signed-distance fields to avoid collisions which can be computed from the included watertight meshes. In other cases such as CBiRRT [58] compute collisions directly using an optimized mesh collision checker.

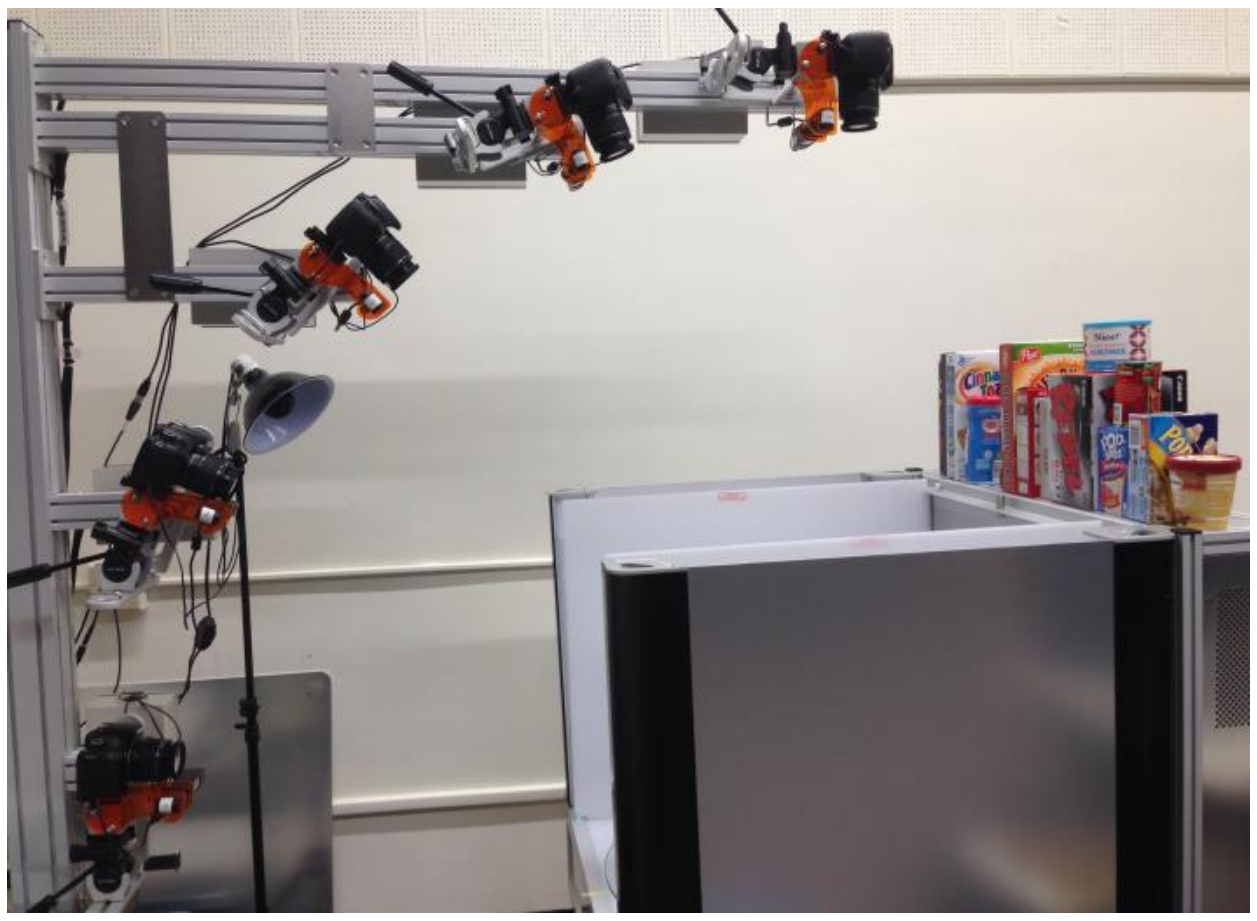

**Fig. 8: BigBIRD Object Scanning Rig: the box contains a computer-controlled turntable.**

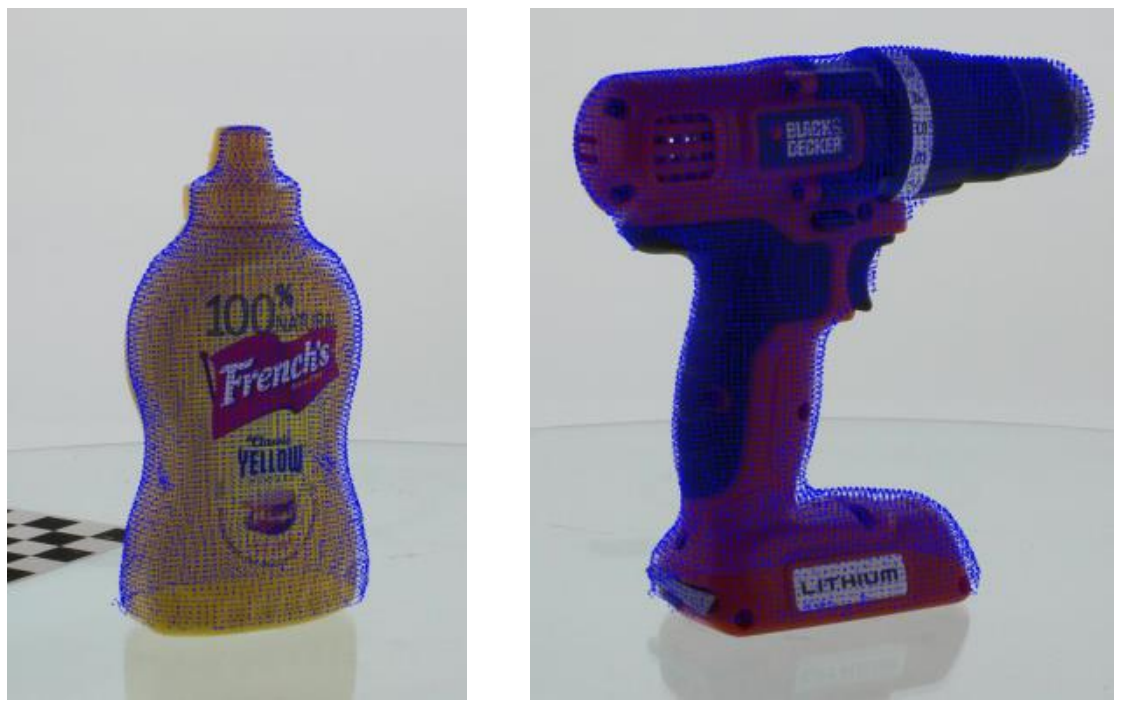

**Fig. 9: Point cloud and textural data overlays on two YCB objects: mustard bottle and power drill.**

In many cases, collision checking is a computational bottleneck for motion planning. Execution time can be reduced using a simplified mesh produced either by hand or with automatic decimation methods [59]. We have not yet provided simplified meshes in this dataset, but view this as an opportunity in future work to further explore mesh approximation algorithms and their impact on motion planning problems using the standardized benchmarks.

### *D. Functional Demonstration*

Fig. 10 demonstrates the entire pipeline. Here, we see the HERB robot [60] preparing to grasp the virtual drill object. This demonstration uses an integration of ROS and OpenRAVE. ROS is used to provide communication between the various hardware and software components of the robot, while OpenRave handles planning and collision checking.

Inside OpenRAVE, the HERB robot uses CBiRRT, the OMPL [61] library and CHOMP to plan and optimize motion trajectories. Using these tools, chains of several actions can be executed in sequence. The simulation environment also provides a mechanism for incorporating feedback from perception systems, which similarly benefit from this dataset. The provided images, meshes and physical objects can all be used as training data for various object detection and pose estimation algorithms, which can then be incorporated into the manipulation pipeline.

Access to both the physical object and a corresponding model for simulation is important for developing and testing new planning and manipulation algorithms. This dataset vastly reduced the time required to set up this example by providing access to objects and meshes that have already been prepared for this purpose. This removed the burden of scanning or modeling new objects and provides benchmark environments that streamline experimental design.

## IV. Protocol Design For Manipulation

A standard set of objects and associated models are a great starting point for common replicable research and benchmarking in manipulation, but there must be a sufficient amount of specification about what should be done with the objects in order to directly compare approaches and results. Given the wide range of technical interests, research approaches, and applications being examined in the manipulation research community, along with how quickly the field moves, we cannot possibly provide sufficient task descriptions that will span the range of interests and remain relevant long-term. Instead, we seek to lay the groundwork for those to be driven by the research community and sub-communities. We therefore focus on two efforts: developing a framework for task protocols, setting, formatting and content guidelines to facilitate effective community-driven specification of standard tasks; and a preliminary set of example protocols that we believe are relevant for our respective communities and approaches, along with experimental implementation of those, including reporting the performance outcomes.

In order to enable effective community-driven evolution of protocols and benchmarks, the web portal associated with this effort will serve as a jumping-off point. Protocols are hosted on arxiv.org, allowing them to be easily posted, shared, and cited, as well as easily updated as the community gives feedback and identifies shortcomings. Our portal will provide links to all protocols that meet the standards laid out in the template, and will provide a forum for discussion on individual protocols.

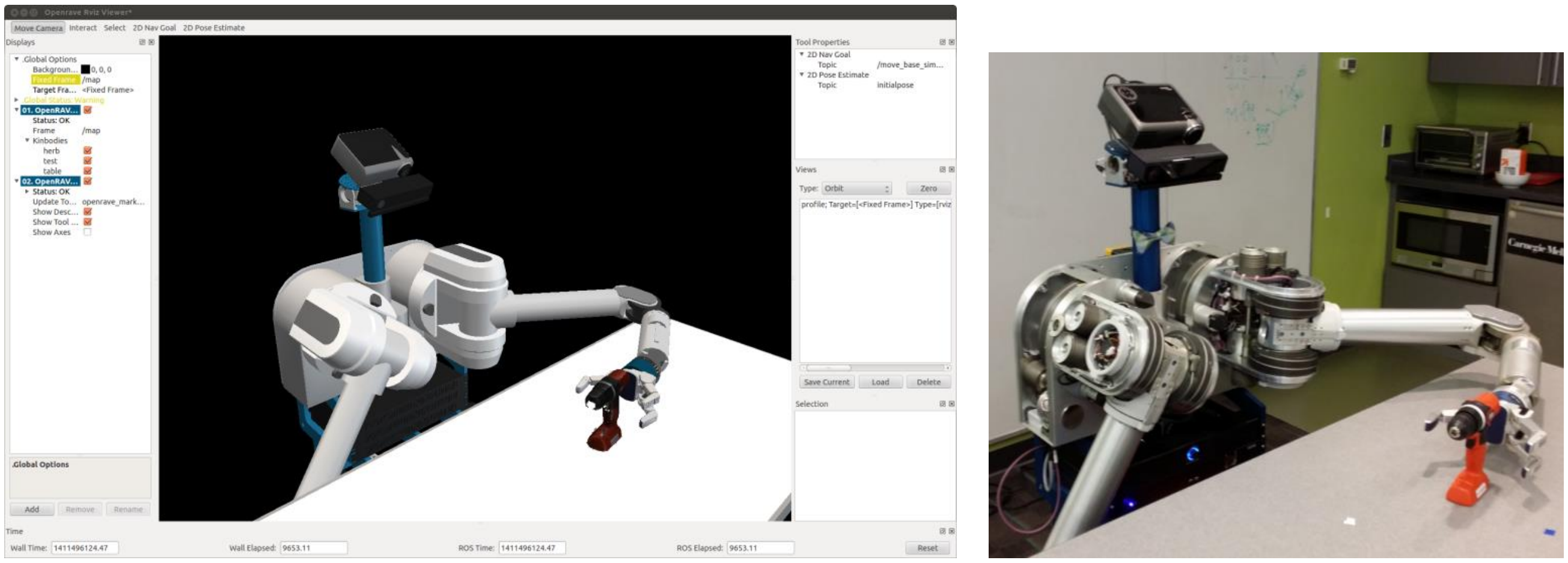

**Fig. 10: (left) Screen-capture from Openrave simulation and planning environment showing the HERB robot [34] planning a grasp of the power drill object in the set. (right) actual grasp being executed by the robot on the physical object.**

### A. *Protocol guidelines*

While developing protocols and benchmarks, one challenging aspect is to decide on the level of detail. Providing only high level descriptions of the experiment, in other words setting too few constraints, makes the repeatability of a benchmark questionable as well as its ability to assess the performance; variations caused by incomplete descriptions of test setups and execution processes induce discrepancy in measurements and won't speak to some quantifiable performance. On the other hand, supplying too many constraints may limit a protocol's applicability, and therefore narrows down its scope. For example, due to the variety of utilized hardware by different research groups in robotics field, satisfying constrained hardware descriptions is not usually possible or preferred.

The aim of this section is to provide guidelines that help to maintain both reliable and widely applicable benchmarks for manipulation. For this purpose, five categories of information are introduced for defining manipulation protocols, namely (1) task description, (2) setup description, (3) robot/hardware/subject description, (4) procedure, and (5) execution constraints. These categories are explained below, and the protocol template provided in Appendix A:

1) *Task description:* Task description is the highest level of information about the protocol. It describes the main action(s) of a task and (most of the time implicitly) the expected outcome(s). In this level, no constraints are given on the setup layout or how the task should be executed. Some task description examples are "pouring liquid from a pitcher to a glass," "hammering a nail on a wood," or "grasping an apple".

2) *Setup Description:* This category provides the list of target objects of the manipulation experiment, their descriptions and initial poses with respect to each other. Also, if there are any other objects used as obstacles or clutter in the manipulation scenario, their description and layout need to be described. For instance, if the task is pouring a liquid from a pitcher to a glass, the object properties of the pitcher and glass should be provided, and their initial poses should be defined. As discussed in the previous sections, the usage of non-standard objects introduces uncertainty to many manipulation experiments presented in literature. We believe that removing uncertainties in this category of information is crucial to maintain well-defined benchmarks. Providing the YCB object and model set is a step towards that purpose. Also, in the protocols proposed in this paper, the initial poses of the objects are accurately provided.
Naturally, a task description can have various setup descriptions designed to assess the manipulation performance in different conditions.

3) *Robot/Hardware/Subject Description:* This category provides information about the task executor. If the protocol is designed for a robotic system, the initial state of the robot with respect to the target object(s) and *a priori* information provided to the robot about the manipulation operation (e.g. the semantic information about the task, whether or not object shape models are provided etc.) are specified in this category. Also, if the protocol is designed for a specific hardware setup (including sensory suite), the description is given. If the task executor is a human subject, how the subject is positioned with respect to the manipulation setup is described here.

*4) Procedure:* In this category, actions that are needed to be taken by the person who conducts the experiment are explained step by step.

*5) Execution Constraints:* In this category, the constraints on how to execute the task are provided. For instance in Box and Blocks Test the subject is expected to use his/her dominant hand, and needs to transfer one block at a time, or if the task is "fetching a mug", the robot may be required to grasp the mug from its handle.

In Appendix A, we provide a template for easily designing manipulation protocols using the abovementioned categories. The amount and specifics of the detail in a specific protocol will naturally vary based on the particular problem being examined, and therefore the insight of the authors about the intended application will be crucial in crafting an effective set of task descriptions and constraints.

Related to this point, we anticipate protocols to be regularly improved and updated with feedback from the research community.

## B. *Benchmark guidelines:*

After the task description, the second major part of each protocol is the specification of the associated benchmark specifying the details of the performance on the protocol. Benchmarks allow the researchers to specify the performance of their system or approach, and enable direct comparison with other approaches. The following categories of information are introduced for defining manipulation benchmarks.

1) *Adopted Protocol:* A well-defined description can be obtained for a manipulation benchmark by adopting a protocol that is defined considering the abovementioned aspects.

2) *Scoring:* Providing descriptive assessment measures is crucial for a benchmark. The output of the benchmark should give reasonable insight of the performance of a system. While designing the scoring criteria, it is usually a good practice to avoid binary (success/fail) measures; if possible, the scoring should include the intermediate steps of the task, giving partial points for a reasonable partial execution.

*3) To Submit:* This field specifies what it is expected from the user of the benchmark. Ideally, the user gives detailed information about anything that is not specified by the protocol (if a robotic manipulation benchmark is considered, these could be robot type, gripper type, grasping strategy, motion planning algorithm, grasp synthesis algorithm, etc.). The user should also submit the resulting score of the test. Moreover, asking the user to submit the reasons of the failed attempts and the factors that bring success would be quite useful for people who analyze the results.

# V. YCB PROTOCOLS AND BENCHMARKS:

While this protocol structure definition (and template provided in Appendix A) helps to guide the development of effective task specification for various manipulation benchmarks, we have developed a number of example protocols to both provide more concrete samples of the types of task definitions that can be put forward as well as specific and useful benchmarks for actually quantifying performance. We have defined five protocols to date:

- Pitcher-Mug Protocol,
- Gripper Assessment Protocol,
- Table Setting Protocol,
- Block Pick and Place Protocol,
- Peg Insertion Learning Assessment Protocol.

From each protocol, a benchmark of reported performance is derived with the same name. We have implemented each of the protocols experimentally and report the benchmark performance of our implementations for each. These protocols and benchmarks can be found in Appendix B and the benchmarking results are presented in Appendix C. *(Note that we intend to move the appendices to arxiv.org after peer review and not have them directly printed with the article.).*

We have also implemented the Box and Blocks Test for maintaining a baseline performance of this test for robotic manipulation.

Short descriptions for the protocol and benchmarks and summary of the benchmarking results are provided below.

### *1) <u>Pitcher-Mug Protocol and Benchmark:</u>*

One of the popular tasks among robotics researchers is pouring a liquid from a container. This task is interesting as it necessitates semantic interpretation, and smooth and precise manipulation of the target object. In Appendix B.1, a protocol is designed for executing this manipulation task. The protocol uses the pitcher and the mug of YCB object and model set, and provides a scenarios by specifying ten initial configurations of the pitcher and the mug. By standardizing the objects and providing detailed initial state information, it is aimed to maintain a common basis of comparison between different research groups. The benchmark derived from this protocol uses a scoring scheme that penalties the amount of liquid that remains in the pitcher or spilled on the table. This benchmark was applied using the HERB robot platform [60] which can be seen in Fig. 11. The reported results show that the task is successfully executed for 8 out of 10 pitcher mug configurations. For the two failed cases, the robot is able to grasp the pitcher, but cannot generate a suitable path for pouring the liquid. This shows the importance of planning the manipulation task as a whole rather than in segments. The details of the experimental results can be seen at Appendix C.1.

### *2) <u>YCB Gripper Assessment Protocol and Benchmark:</u>*

The abilities of a robot's gripper affect its manipulation performance significantly. In literature and in commercial market, various gripper designs are available each of which have different manipulation capabilities. The protocol presented in Appendix B.2 defines a test procedure for assessing the performance of grippers for grasping objects of various shapes and sizes. This protocol utilizes objects from the shape and tool categories of the YCB object and model set. Using this protocol, a benchmark is defined based on a scoring table. We applied this benchmark to two grippers designed in Yale GRAB Lab, the Model T and Model T42 [62], which can be seen in Fig. 12. The results show that the Model T can provide successful grasp for only a limited range of object sizes. This gripper is not suitable for grasping small and flat object. However, the ability to interlace its fingers increases the contact surface with the object and brings an advantage especially for grasping concave and articulated objects. The Model T42 is able to provide stable power grasps for large objects and precision grasps for small objects. This model is also successful in grasping flat objects thanks to its nail-like finger tips. However, not being able to interlace its fingers brings a disadvantage while grasping articulated objects. Using the same benchmark for evaluating different gripper designs did not only provide a basis of comparison, but also gave many clues about how to improve the designs. The details of the experimental results are presented in Appendix C.2.

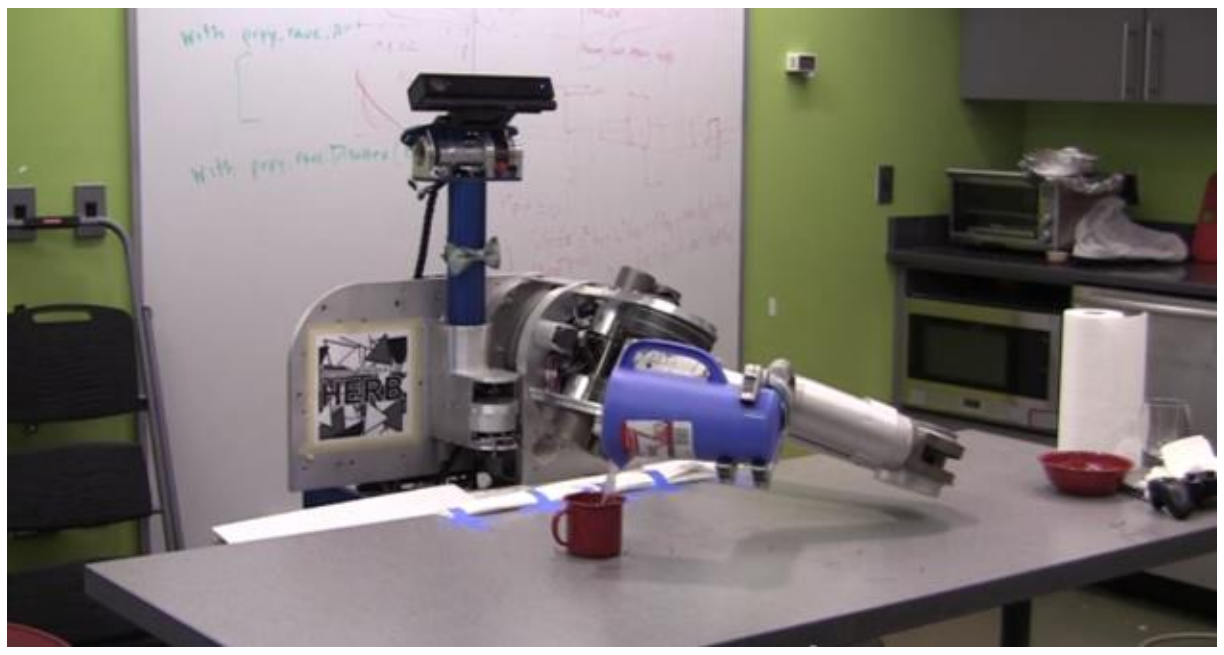

**Fig. 11: HERB robot implementing Pitcher-Mug Benchmark**

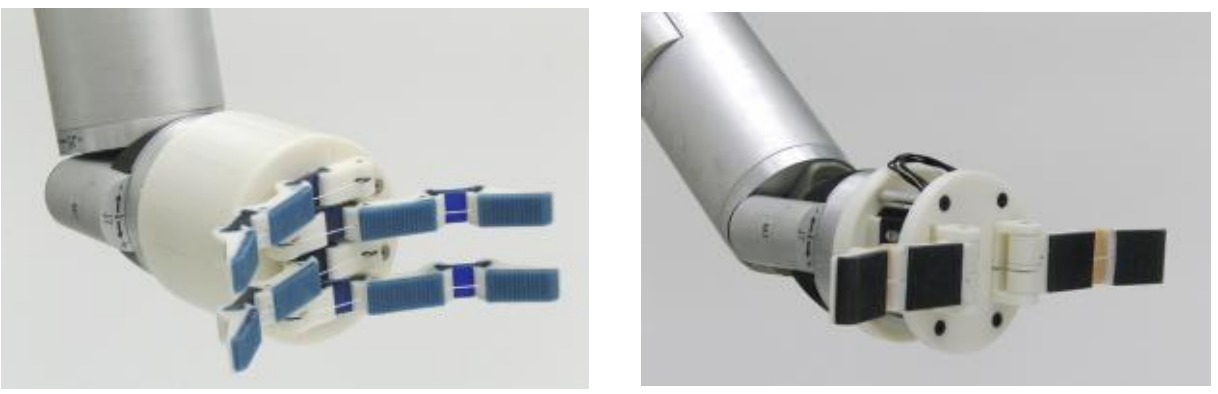

(a) (b)

**Fig. 12: Grippers compared with Gripper Assessment Benchmark. (a) Model T, (b) Model T42.**

*3) YCB Protocol and Benchmark for Table Setting:*

Pick-and-place is an essential ability for service robots. The benchmark provided in Appendix B.3 assesses this ability by the daily task of table setting. The protocol uses the mug, fork, knife, spoon, bowl and plate of the YCB object and model set. These objects are placed to predefined initial locations, and the robot is expected to replace them to specific final configurations. The benchmark scores the performance of the robot by the accuracy of the final object poses. This benchmark can also be applied in a simulation environment since the models of the objects are provided by the YCB Object and Model Set. A URDF file which spawns the scenario for Gazebo simulation environment is given at *http://rll.eecs.berkeley.edu/ycb/*. A snapshot of this setting can be seen in Fig. 13.

*4) Block Pick and Place Protocol and Benchmark:*

Manual dexterity and the manipulation of small objects are critical skills for robots in several contexts. The block pick and place protocol presented in Appendix B.4 is designed to test a robot's ability to grasp small objects and transfer them to a specified location. This task is an important test of both arm and gripper hardware and motion planning software, as both contribute to overall dexterity. Points are awarded based on completion and precision of the manipulation. We executed this test on the HERB robot [60] as seen in Fig. 14. An image of the printed layout with the placed blocks after task completion can be seen in Fig. 15. The results show that the robot is not able to success in precise pick and place task. The main reason is the utilized open loop grasping approach: The robot executes a robust push grasp strategy which allows it to grasp the blocks successfully. However, the pose of the block with respect to the gripper is not known precisely after the grasp. This prevents placing the blocks accurately to the target locations. The details of the experimental results are presented in Appendix C.3.

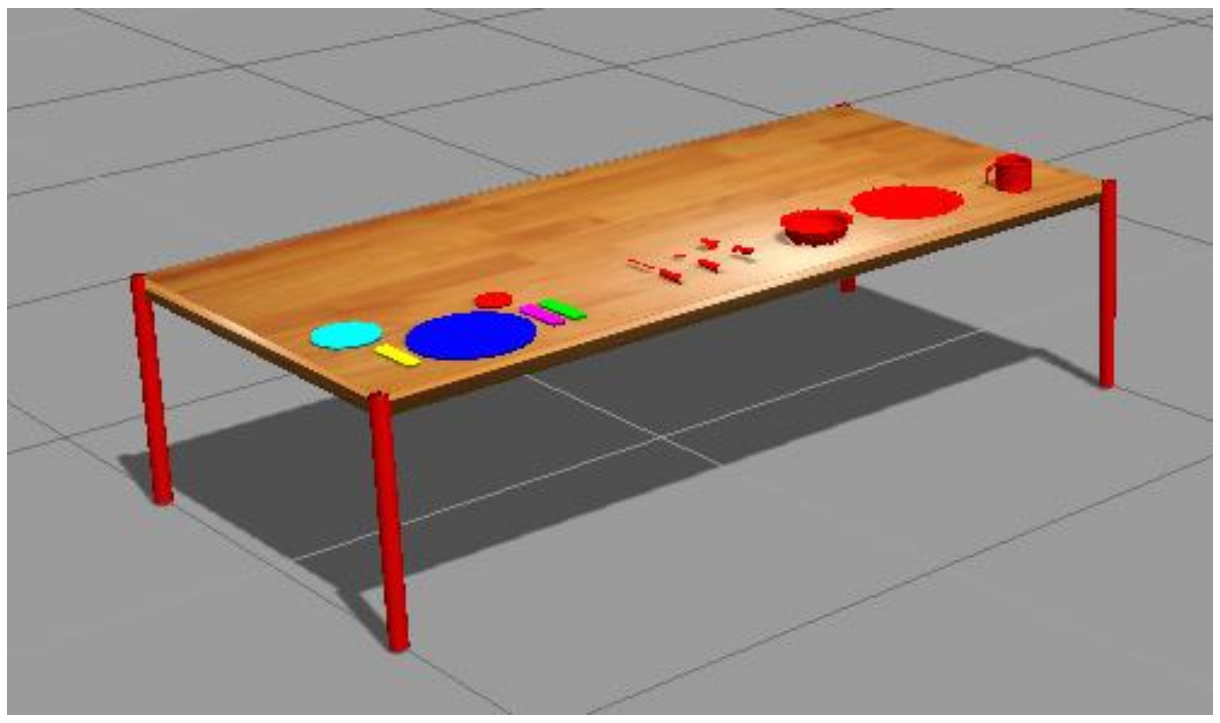

**Fig. 13: The simulation environment for Table Setting Benchmark. This environment can be spawned by using the URDF provided at http://rll.eecs.berkeley.edu/ycb**

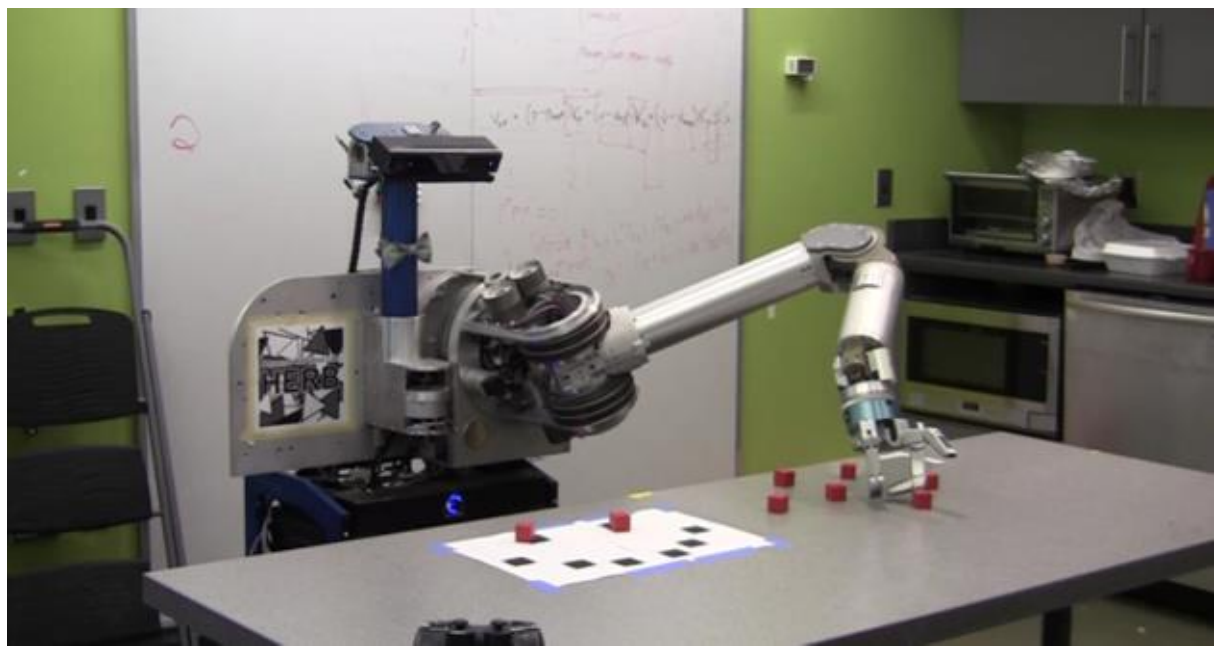

**Fig. 14: HERB robot implementing Block Pick and Place Benchmark**

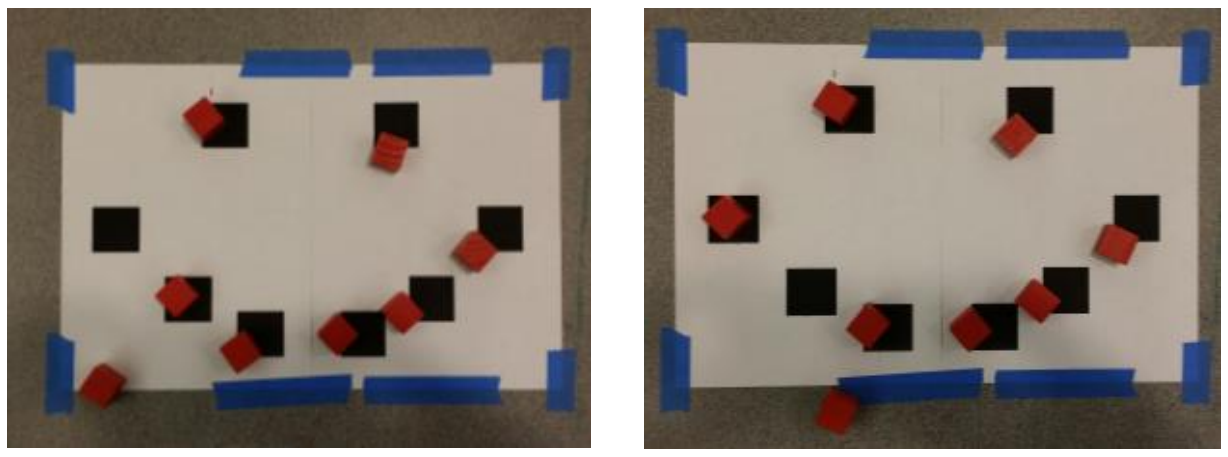

**Fig. 15: The results of the Block Pick and Place Benchmark**

### 5) *Peg Insertion Learning Assessment Protocol and Benchmark:*

The Peg Insertion Learning Assessment Benchmark is designed for allowing comparison between various learning techniques. The benchmark measures the performance of a learned peg insertion action under various positioning perturbations. The perturbations are applied by moving the peg board to a random direction for certain amount of distance. We applied this benchmark to assess the performance of a learned linear-Gaussian controller using a PR2 robot [63] (Fig. 16). The state of the controller consists of the joint angles and angular velocities of the robot, and the positions and velocities of three points in the space of the end effector (3 points in order to fully define a rigid body configuration). No information is available to the controller at run time except for this state information. The results show that, the learned controller shows reasonable performance, 4 success out of 10 trials, for the case of 5mm position perturbation to a random direction. This success rate can be achieved by executing the controller for only one second. However, the performance does not improve even if the controller is run for a longer period of time. In the case of 10mm position perturbation, the controller fails completely. We are planning to learn the same task with different learning techniques and compare their performances using the benchmark.

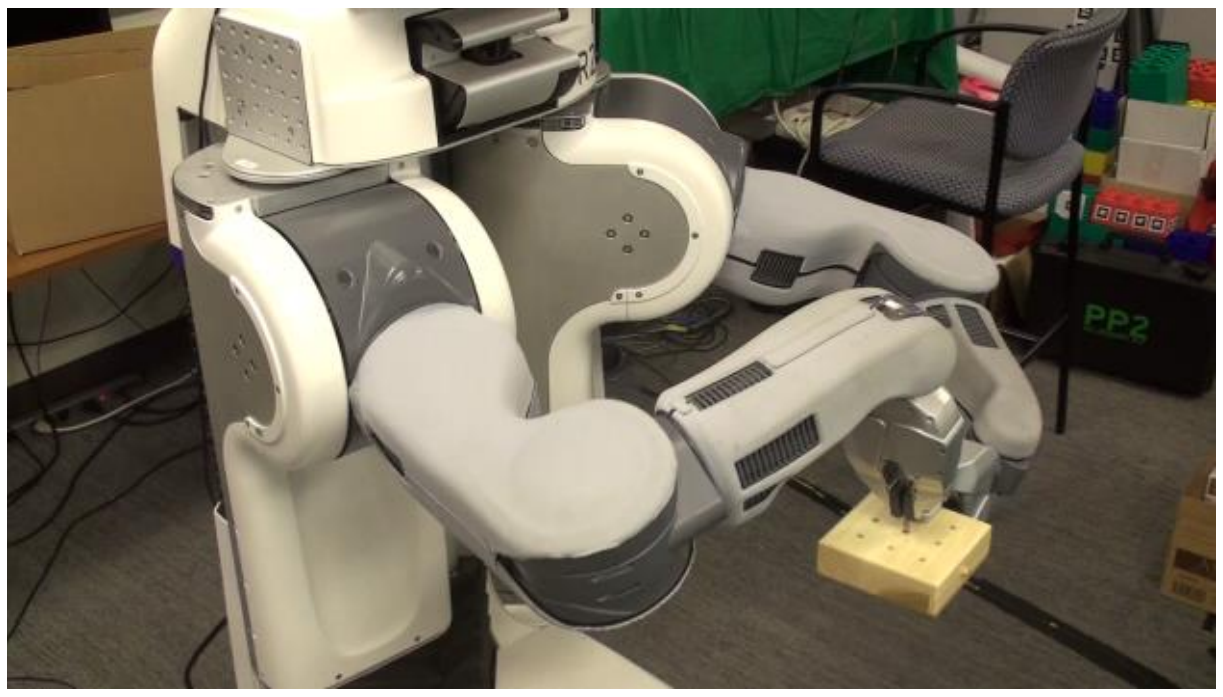

**Fig. 16: PR2 executing the Peg Insertion Learning Assessment Benchmark.**

### 6) *Box and Blocks Test:*

As mentioned previously in Section 2, the Box and Blocks Test [41] is a widely used assessment technique that is utilized in prosthetics and rehabilitation fields. The test evaluates how many blocks can be grasped and moved from one side of the box (Fig. 17) to the other in a fixed amount of time. We believe that the application of this test can also be quite useful for assessing the manipulation capabilities of robots. In order to establish a baseline performance for this test for robotic manipulators, we applied the Box and Blocks Test with a PR2 robot (Fig. 17) by implementing a very simple heuristic rules: The robot picks a location from a uniform distribution over the box and attempts to pick up a block. The gripper's pose aligns with the length of the box. The gripper is then closed, and checked if it is fully closed. If the gripper closes fully, this means no blocks have been grasped and therefore the robot chooses a new location to attempt another pick. The robot repeats this heuristic until the gripper is not fully closed. When a grasp is detected, the robot moves to the destination box and releases the block. By using this heuristic, we run 10 experiments of 2 minutes each, and report the results in Appendix C.5.

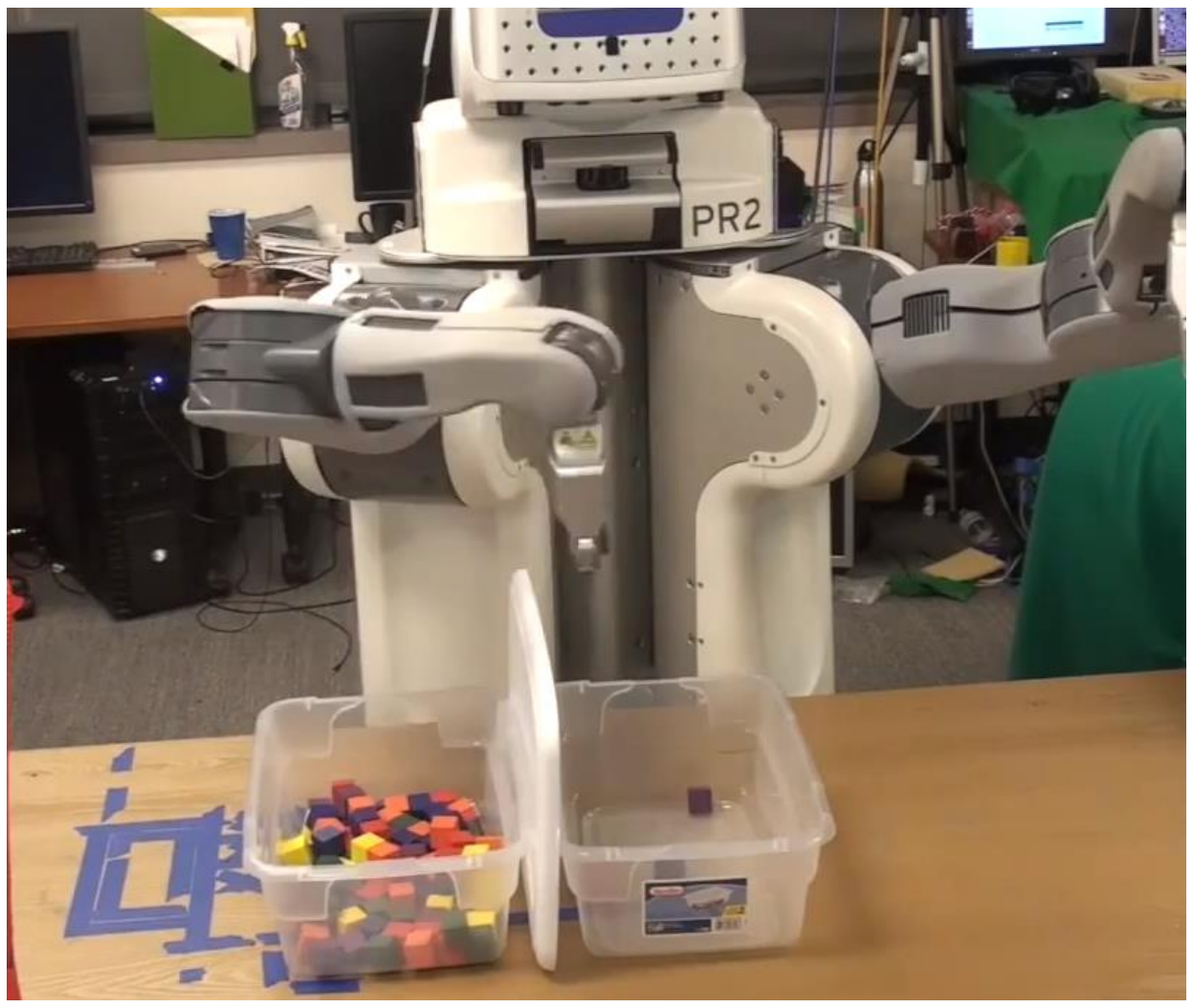

**Fig. 17: PR2 executing the Box and Blocks Test.**

## VI. Conclusions and Future Work

This paper proposes a set of objects and related tasks, as well as high-resolution scans and models of those objects, intended to serve as a widely-distributed and widely-utilized set of standard objects to facilitate the implementation of standard performance benchmarks for robotic grasping and manipulation research. The objects were chosen based on an in-depth literature review of other objects and tasks previously proposed and utilized in robotics research, with additional consideration to efforts in prosthetics and rehabilitation. Furthermore, a number of practical constraints were considered, including a reasonable total size and mass of the set for portability, low cost, durability of the objects, and the likelihood that the objects would remain mostly unchanged in years to come. High-resolution RGBD scans of the objects were done and models of the objects have been constructed to allow easy portability into simulation and planning environments. All of these data are freely available in the associated repository [54]. The objects sets will be freely distributed to a large number of research groups through workshops/tutorials associated with this effort, and will be made available to purchase otherwise.

While a common set of widely-available objects is a much-needed contribution to the manipulation research community, the objects themselves are just the beginning. The generation of appropriate detailed tasks and protocols involving the objects is ultimately what will allow for replicable research and performance comparison. We make inroads into that problem in this paper by proposing a structure for protocols and benchmarks, implemented in a template, as well as six example protocols. The specification of protocols and benchmarks will necessarily need to be sub-community driven and continually evolving – specific aspects of the manipulation and specific research questions and interests will naturally require different task particulars (e.g. specified and free parameters). We therefore plan to involve the research community in this effort via a web portal and arXiv-style working documents for proposed protocols, and will work towards having the majority of those protocols come from the user community rather than the authors. Additionally, we plan to have on this portal a "records" keeping functionality to keep track of the current "world records" for the different tasks and protocols, along with video and detailed descriptions of the approaches utilized, generating excitement, buzz, motivation, and inspiration for the manipulation community to compare approaches and push forward the state of the art.

Other efforts that we plan to undertake include more detail about the objects proposed, including information about the inertia of the objects, as well as frictional properties between the objects and common surfaced. Additionally, we will expand our treatment of the modelling of the objects, including addressing the tradeoffs between number of "triangles" and the reliable representation of the object geometry. Furthermore, before final publication and distribution of the object set, we will seek additional input from the research community on the specific objects in the set.

It is our hope that this work will help to address the long-standing need for common performance comparisons and benchmarks in the research community and will provide a starting point for further focused discussion and iterations on the topic.

## Acknowledgment

The authors would like to thank Michael Leddy for his efforts in measuring the physical properties of the objects in the set. We would also like to thank Peter Allen and Jeff Trinkle for their feedback on the objects in the set and the overall approach taken here.

## References

[1] A. Singh, J. Sha, K. S. Narayan, T. Achim, and P. Abbeel, "BigBIRD: A Large-Scale 3D Database of Object Instances," in *International Conference on Robotics and Automation*, 2014.

[2] B. Li, Y. Lu, C. Li, A. Godil, T. Schreck, M. Aono, *et al.* (2014). *SHREC'14 Track: Large Scale Comprehensive 3D Shape Retrieval*. Available: http://www.itl.nist.gov/iad/vug/sharp/contest/2014/Generic3D/

[3] C. Goldfeder, M. Ciocarlie, H. Dang, and P. K. Allen, "The Columbia Grasp Database," in *Robotics and Automation, 2009. ICRA '09. IEEE International Conference on*, 2009, pp. 1710-1716.

[4] (2014). *Sketchup*. Available: http://www.sketchup.com/

[5] S. Chitta, I. Sucan, and S. Cousins, "MoveIt!," *IEEE Robotics Automation Magazine*, vol. 19, pp. 18-19, March 2012.

[6] Y. S. Choi, T. Deyle, T. Chen, J. D. Glass, and C. C. Kemp, "A list of household objects for robotic retrieval prioritized by people with ALS," in *Rehabilitation Robotics, 2009. ICORR 2009. IEEE International Conference on*, 2009, pp. 510-517.

[7] http://amazonpickingchallenge.org/.

[8] J. Adams, K. Hodges, J. Kujawa, and C. Metcalf, "Test-retest Reliability of the Southampton Hand Assessment Procedure," *International Journal of Rehabilitation Research*, vol. 32, p. S18, 2009.

[9] S. Kalsi-Ryan, A. Curt, M. Verrier, and F. MG, "Development of the Graded Redefined Assessment of Strength, Sensibility and Prehension (GRASSP): reviewing measurement specific to the upper limb in tetraplegia," *Journal of Neurosurg Spine*, vol. 1, pp. 65-76, 2012.

[10] N. Yozbatiran, L. Der-Yeghiaian, and C. SC, "A standardized approach to performing the action research arm test," *Neurorehabil Neural Repair*, vol. 1, pp. 78-90, 2008.

[11] E. D. Sears and K. C. Chung, "Validity and responsiveness of the Jebsen-Taylor Hand Function Test," *J Hand Surg Am*, vol. 35, pp. 30-7, Jan 2010.

[12] A. P. del Pobil, "Benchmarks in Robotics Research," ed, 2006.

[13] R. Madhavan, R. Lakaemper, and T. Kalmar-Nagy, "Benchmarking and standardization of intelligent robotic systems," in *Advanced Robotics, 2009. ICAR 2009. International Conference on*, 2009, pp. 1-7.

[14] I. Iossifidis, G. Lawitzky, S. Knoop, and R. Zöllner, "Towards Benchmarking of Domestic Robotic Assistants," in *Advances in Human-Robot Interaction*. vol. 14, ed: Springer Berlin Heidelberg, 2005, pp. 403-414.

[15] R. Detry, O. Kroemer, and D. Kragic, "International Workshop on Autonomous Grasping and Manipulation: An Open Challenge," ed, 2014.

[16] (2006). *AIM@SHAPE Shape Repository v4.0*. Available: http://shapes.aim-at-shape.net/viewmodels.php

[17] A. Richtsfeld. (2012). *The Object Segmentation Database (OSD)*. Available: http://www.acin.tuwien.ac.at/?id=289

[18] P. Shilane, P. Min, M. Kazhdan, and T. Funkhouser, "The Princeton Shape Benchmark," in *Shape Modeling Applications, 2004. Proceedings*, 2004, pp. 167-178.

[19] A. Kasper, Z. Xue, and R. Dillmann, "The KIT object models database: An object model database for object recognition, localization and manipulation in service robotics," *The International Journal of Robotics Research*, 2012.

[20] S. Ulbrich, D. Kappler, T. Asfour, N. Vahrenkamp, A. Bierbaum, M. Przybylski, *et al.*, "The OpenGRASP benchmarking suite: An environment for the comparative analysis of grasping and dexterous manipulation," in *Intelligent Robots and Systems (IROS), 2011 IEEE/RSJ International Conference on*, 2011, pp. 1761-1767.

[21] G. Kootstra, M. Popovic, J. A. Jorgensen, D. Kragic, H. G. Petersen, and N. Kruger, "VisGraB: A benchmark for vision-based grasping," *Paladyn*, vol. 3, pp. 54-62, 2012.

[22] G. Kragten, A. C. Kool, and J. Herder, "Ability to hold grasped objects by underactuated hands: Performance prediction and experiments," in *Robotics and Automation, 2009. ICRA '09. IEEE International Conference on*, 2009, pp. 2493-2498.

[23] G. A. Kragten, C. Meijneke, and J. L. Herder, "A proposal for benchmark tests for underactuated or compliant hands," *Mechanical Sciences*, vol. 1, pp. 13-18, 2010.

[24] B. Li, A. Godil, M. Aono, X. Bai, T. Furuya, L. Li, *et al.* (2012). *SHREC'12 Track: Generic 3D Shape Retrieval*. Available: http://www.itl.nist.gov/iad/vug/sharp/contest/2012/Generic3D/

[25] A. Tatsuma, H. Koyanagi, and M. Aono, "A large-scale Shape Benchmark for 3D object retrieval: Toyohashi shape benchmark," in *Signal Information Processing Association Annual Summit and Conference (APSIPA ASC), 2012 Asia-Pacific*, 2012, pp. 1-10.

[26] R. B. Rusu. (2011). *NIST and Willow Garage: Solutions in Perception Challenge*. Available: http://www.nist.gov/el/isd/challenge-022511.cfm; http://www.willowgarage.com/blog/2011/02/28/nist-and-willow-garage-solutions-perception-challenge

[27] H. Dutagaci, A. Godil, P. Daras, A. Axenopoulos, G. Litos, S. Manolopoulou, *et al.* (2011). *SHREC'11 Track: Generic Shape Retrieval*. Available: http://www.itl.nist.gov/iad/vug/sharp/contest/2011/NonRigid/

[28] A. Janoch, S. Karayev, Y. Jia, J. T. Barron, M. Fritz, K. Saenko, *et al.*, "A category-level 3-D object dataset: Putting the Kinect to work," in *Computer Vision Workshops (ICCV Workshops), 2011 IEEE International Conference on*, 2011, pp. 1168-1174.

[29] K. Lai, L. Bo, X. Ren, and D. Fox, "A large-scale hierarchical multi-view RGB-D object dataset," in *Robotics and Automation (ICRA), 2011 IEEE International Conference on*, 2011, pp. 1817-1824.

[30] Z. Lian, A. Godil, T. Fabry, T. Furuya, J. Hermans, R. Ohbuchi, *et al.* (2010). *SHREC'10 Track: Non-rigid 3D Shape Retrieval*. Available: http://tosca.cs.technion.ac.il/book/shrec_robustness2010.html

[31] R. Wessel, m. Bl\"u, Ina, and R. Klein, "A 3D Shape Benchmark for Retrieval and Automatic Classification of Architectural Data," in

*Eurographics 2009 Workshop on 3D Object Retrieval*, 2009, pp. 53-56.
[32] N. Iyer, S. Jayanti, and K. Ramani, "An Engineering Shape Benchmark for 3D Models," in *Proceedings of IDETC/CIE 2005 ASME 2005 International Design Engineering Technical Conferences \& Computers and Information in Engineering Conference*, 2005.
[33] R. Fang, A. Godil, X. Li, and A. Wagan, "A New Shape Benchmark for 3D Object Retrieval," in *Proceedings of the 4th International Symposium on Advances in Visual Computing*, Berlin, Heidelberg, 2008, pp. 381-392.
[34] (2005). *McGill 3D Shape Benchmark*. Available: http://www.cim.mcgill.ca/~shape/benchMark/
[35] N. Kapadia, V. Zivanovic, M. Verrier, and M. R. Popovic, "Toronto Rehabilitation Institute-Hand Function Test: Assessment of Gross Motor Function in Individuals With Spinal Cord Injury," in *Topics in Spinal Cord Injury Rehabilitation*, 2012.
[36] R. W. Sumner and J. Popovic, "Deformation Transfer for Triangle Meshes," *ACM Transactions on Graphics,* vol. 3, 2014.
[37] (2003). *NTU 3D Model Benchmark*. Available: http://3d.csie.ntu.edu.tw/
[38] (2014). *the ITI database*. Available: http://vcl.iti.gr/3d-object-retrieval/
[39] (2014). *Model Bank Library*. Available: http://digimation.com/3d-libraries/model-bank-library/
[40] T. Wisspeintner, T. van der Zan, L. Iocchi, and S. Schiffer, "RoboCup@Home: Results in Benchmarking Domestic Service Robots," in *RoboCup 2009: Robot Soccer World Cup XIII*. vol. 5949, ed: Springer Berlin Heidelberg, 2010, pp. 390-401.
[41] V. Mathiowetz, G. Volland, N. Kashman, and K. Wever, "Adult norms for the box and blocks test of manual dexterity," *The American journal of occupational therapy,* vol. 89, pp. 386-391, 1985.
[42] V. Mathiowetz, K. Weber, N. Kashman, and G. Volland, "Adult norms for the Nine Hole Peg Test of finger dexterity," *Occupational Therapy Journal of Research,* vol. 5, pp. 24-38, 1985.
[43] S. Rehab. *http://saliarehab.com/*.
[44] *GRASSP*. Available: grassptest.com
[45] *Grasp and Release Test*. Available: http://www.rehabmeasures.org/Lists/RehabMeasures/DispForm.aspx?ID=1053
[46] L. Resnik, L. Adams, M. Borgia, J. Delikat, R. Disla, C. Ebner*, et al.*, "Development and Evaluation of the Activities Measure for Upper Limb Amputees," *Archives of Physical Medicine and Rehabilitation,* vol. e4, pp. 488-494, 2013.
[47] M. Popovic and C. Contway, "Rehabilitation Engineering Laboratory Hand Function Test For Functional Electrical Stimulation Assisted Grasping," ed, 2003.
[48] H. Burger, F. Franchignoni, A. W. Heinemann, S. Kotnik, and A. Giordano, "Validation of the Orthotics and Prosthetics User Survey Upper Extremity Functional Status Module in People with Unilateral Upper Limb Amputation," *Journal of Rehabilitation M,* vol. 40, pp. 393-399, 2008.
[49] M. R. Popovic, T. Keller, I. P. I. Papas, V. Dietz, and M. Morari, "Surface-stimulation technology for grasping and walking neuroprostheses," *IEEE Engineering in Medicine and Biology Magazine,* vol. 1, pp. 82-93, 2001.
[50] J. Harris, J. Eng, W. Miller, and D. AS, "A self-administered Graded Repetitive Arm Supplementary Program (GRASP) improves arm function during inpatient stroke rehabilitation: a multi-site randomized controlled trial," *Stroke,* vol. 40, pp. 2123-2131, 2009.
[51] K. A. Stubblefield, L. A. Milller, R. D. Lipschutz, M. E. Phillips, C. W. Heckathorne, and T. A. Kuiken, "Occupational Therapy Outcomes with Targeted Hyper-Reinnervation Nerve Transfer Surgery: Two Case Studies," in *Proceedings of the MyoElectric Controls/Powered Prosthetics Symposium*, 2005.
[52] K. Matheus and A. M. Dollar, "Benchmarking grasping and manipulation: Properties of the Objects of Daily Living," in *IROS*, 2010, pp. 5020-5027.
[53] *The Dash - "Disabilities of the Arm, Shoulder and Hand"*. Available: http://dash.iwh.on.ca/system/files/dash_questionnaire_2010.pdf
[54] *http://rll.eecs.berkeley.edu/ycb/*.
[55] M. Quigley, K. Conley, B. Gerkey, J. Faust, T. Foote, J. Leibs*, et al.*, "ROS: an open-source Robot Operating System," *ICRA workshop on open source software,* vol. 3, p. 5, 2009.
[56] R. Diankov and J. Kuffner, "OpenRAVE: A Planning Architecture for Autonomous Robotics," 2008.
[57] N. Ratliff, M. Zucker, J. A. Bagnell, and S. Srinivasa, "CHOMP: Gradient Optimization Techniques for Efficient Motion Planning," *Icra: 2009 Ieee International Conference on Robotics and Automation, Vols 1-7,* pp. 4030-4035, 2009.
[58] D. Berenson, S. S. Srinivasa, D. Ferguson, and J. J. Kuffner, "Manipulation planning on constraint manifolds," in *Robotics and Automation, 2009. ICRA '09. IEEE International Conference on*, 2009, pp. 625-632.
[59] M. Garland and P. S. Heckbert, "Surface simplification using quadric error metrics," presented at the Proceedings of the 24th annual conference on Computer graphics and interactive techniques, 1997.
[60] S. S. Srinivasa, D. Berenson, M. Cakmak, A. Collet, M. R. Dogar, A. D. Dragan*, et al.*, "Herb 2.0: Lessons Learned From Developing a Mobile Manipulator for the Home," *Proceedings of the IEEE,* vol. 100, pp. 2410-2428, 2012.
[61] I. A. Sucan, M. Moll, and L. E. Kavraki, "The Open Motion Planning Library," *Ieee Robotics & Automation Magazine,* vol. 19, pp. 72-82, Dec 2012.
[62] R. M. Raymond, U. O. Lael, and A. M. Dollar, "A Modular, Open-Source 3D Printed Underactuated Hands,," in *IEEE International Conference on Robotics and Automation (ICRA)*, Karlsruhe, Germany, 2013.
[63] W. Garage. (2009). *Overview of the PR2 robot*. Available: http://www.willowgarage.com/pages/pr2/overview

### Appendix A. Protocol and Benchmark Template for Manipulation Research:

# MANIPULATION PROTOCOL TEMPLATE

| | |
|---|---|
| Reference No / Version | |
| Authors | |
| Institution | |
| Contact information | |
| Purpose | |
| Task Description | |
| Setup Description | **Description of the manipulation environment:** |
| | **List of objects and their descriptions:** |
| | **Initial poses of the objects:** |
| Robot/Hardware/Subject Description | **Targeted robots/hardware/subjects:** |
| | **Initial state of the robot/hardware/subject with respect to the setup:** |
| | **Prior information provided to the robot/hardware/subject:** |
| Procedure | |
| Execution Constraints | |

# Manipulation Benchmark Template

| | |
|---|---|
| Reference No / Version | |
| Authors | |
| Institution | |
| Contact information | |
| Adopted Protocol | |
| Scoring | |
| To Submit | |

## Appendix B. Protocols and benchmarks

### Appendix B.1 Pitcher-Mug Protocol and Benchmark:

# Pitcher-Mug Protocol

| | |
|---|---|
| Reference No / Version | P-PM-0.01 |
| Authors | Aaron Walsman, Berk Calli |
| Institution | CMU, Yale University |
| Contact information | aaronwalsman@gmail.com, berk.calli@yale.edu |
| Purpose | To assess the ability of a robot to execute a common daily task. |
| Task Description | Pour a liquid from the pitcher to the mug. |
| Setup Description | **List of objects and their descriptions:**<br>The following objects from the YCB object set are used: The mug, the pitcher and the white table cloth. |
| | **Initial poses of the objects:**<br>The initial poses of the mug and the pitcher are defined for ten different scenarios. Printable layouts provided are provided for all these configurations are in http://rll.eecs.berkeley.edu/ycb/. Print the layouts to an A3 paper, and place them on the table one by one for each scenario. |
| | **Description of the manipulation environment:**<br>Conduct the experiment on the white table cloth provided by the object set. Avoid background clutter. Instead of water, rice can be used to avoid hazardous conditions. |
| Robot/Hardware Description | **Targeted robots:**<br>The protocol is designed for robots that have on board sensors. No sensors fixed to the environment should be used. |
| | **Initial state of the robot/subject with respect to the setup:**<br>In the initial state, the robot's vision sensor should be aligned with the corresponding marker on the printable layouts. |
| | **Prior information provided to the robot:**<br>The robot does not know the model of the pitcher or the mug.<br>The robot knows:<br>• both the pitcher and the mug have a handle,<br>• the pitcher's handle is bigger than the mug's handle,<br>• the opening of the pitcher is at the top and opposite side of the handle. |
| Procedure | **1.** Fill the mug with rice or water.<br>**2.** Transfer the rice or water to the pitcher.<br>**3.** Place the next A3 layout to the table. Place the pitcher and the mug on the corresponding places of the printable layout.<br>**4.** Run the system.<br>**5.** Empty the pitcher and the mug.<br>**6.** Repeat steps 1 to 5 for all ten scenarios. |
| Execution Constraints | Same strategy should be used for each layout. |

# Pitcher-Mug Benchmark

| | |
|---|---|
| Reference No / Version | B-PM-0.01 |
| Authors | Aaron Walsman, Berk Calli |
| Institution | CMU, Yale University |
| Contact information | aaronwalsman@gmail.com, berk.calli@yale.edu |
| Adopted Protocol | Pitcher-Mug Protocol (Reference No: P-PM-0.01) |
| Scoring | • At least half of the mug should be filled. Otherwise, the system gets 0 points.<br>• It is aimed to pour the water without spilling it to the table. The following calculation gives 1 point for the perfectly executed task and applies penalties for the spilled water.<br>  ○ Measure the weight of the mug and the pitcher before the execution ($M_{before}$ and $P_{before}$ respectively).<br>  ○ Execute the task.<br>  ○ Measure the weight of the mug and pitcher after the execution ($M_{after}$ and $P_{after}$ respectively).<br>  ○ Calculate $(M_{after}-M_{before})/(P_{before}-P_{after})$.<br>• Implement the protocols for all ten printable layouts and sum up the points for each execution for the overall score. |
| To Submit | • Scores for individual scenarios, and overall score.<br>• Detailed description of the system.<br>• Detailed comments on:<br>  ○ What makes the system successful?<br>  ○ What makes the system fail? |

APPENDIX B.2 GRIPPER ASSESSMENT PROTOCOL AND BENCHMARK:

# Gripper Assessment Protocol

| | |
|---|---|
| Reference No / Version | P-GA-0.01 |
| Authors | Berk Calli, Aaron Dollar |
| Institution | Yale University |
| Contact information | berk.calli@yale.edu |
| Purpose | Assessing the abilities of a gripper to firmly grasp objects of different sizes and shapes. |
| Task Description | Grasp objects with various sizes and shapes one by one. |
| Setup Description | **<u>List of objects and their descriptions:</u>**<br>The following objects from the YCB object and model set are used:<br>• Round objects: Soccer ball, softball, tennis ball, racquetball, golf ball, marbles (4 different sizes).<br>• Flat objects: Washers (7 different sizes.), credit card.<br>• Tools: Pen, scissors, screwdriver, driller, hammer, clamps (4 different sizes).<br>• Articulated objects: Plastic chain, rope.<br>Other objects: a 1 cm thick planar object (not included in the set, for applying position offset to the objects as explained below).<br>The user of the protocol may prefer to use any combination of the above mentioned object categories depending on the scope of the performance assessment. |
| | **<u>Description of the manipulation environment:</u>**<br>Print the grid provided at http://rll.eecs.berkeley.edu/ycb/ and place it on a table. This grid is used to easily apply position offsets to the object in Cartesian directions.<br>Print another copy of the grid and stick it on the 1 cm thick planar object as explained at http://rll.eecs.berkeley.edu/ycb/. Place the planar object on the table by aligning the grids. |
| | **<u>Initial poses of the objects:</u>**<br>In order to assess the robustness of a gripper, position offsets will be applied to the objects while keeping the gripper position fixed. For this purpose, four set points (SP) are defined as follows:<br>• SP 1: grid on the planar object, x = 0 cm, y = 0 cm,<br>• SP 2: grid on the planar object, x = 1 cm, y = 0 cm,<br>• SP 3: grid on the planar object, x = 0 cm, y = 1 cm,<br>• SP 4: grid on the table, x = 0 cm, y = 0cm.<br>For round objects: Four initial positions are defined by placing the objects at SP1, SP2, SP3 and SP4. (To stabilize the round objects, they can be placed on the washers).<br>For flat objects: Three initial positions are defined for each object by aligning the center of the objects with SP1, SP2 and SP3. Set the orientation of the credit card such that the long edge is parallel to the x axis of the grid.<br>For tools: Four initial positions are defined by aligning the (approximate) center of the objects with SP1, SP2, SP3, SP4. Set their orientation such that the major axis is parallel to the x axis of the grid.<br>Articulated objects: Place the objects randomly inside the grid area. |
| Robot/Hardware Description | **<u>Targeted robots/hardware:</u>**<br>Any. |
| | **<u>Initial state of the robot with respect to the setup:</u>**<br>For each object in the round, flat, and tools categories, any desired grasping pose can be chosen when the object is at SP1. This position should be kept the same while conducting experiments with SP2, SP3 and SP4 (see the procedure part for details.) |
| | **<u>Prior information provided to the robot:</u>**<br>None. |
| Procedure | For each object in the round, flat, and tools categories,<br>1. Place the object to SP1.<br>2. Set a desired grasping pose of the gripper, and record this pose.<br>3. Execute the grasp.<br>4. Lift the object.<br>5. Hold for 3 seconds.<br>6. Rotate the grasped object 90 degrees in the x direction of the grid.<br>7. Hold for 3 seconds.<br>8. Move the gripper to the pose recorded at step 2, and repeat the steps 3 to 7 for set points SP2, SP3 and if applicable SP4.<br>If the grasp fails at step 3, 4, 5, 6 or 7 when the object is at SP1, skip to the next object. If the grasp fails when the object is at SP2, SP3 or SP4, jump to step 8.<br>For articulated objects:<br>1. Place the object randomly inside the grid.<br>2. Set a desired grasping pose of the gripper, and record this pose.<br>3. Execute the grasp.<br>4. Lift the object 15 cms.<br>5. Hold for 3 seconds.<br>6. Keeping the recorded pose of the gripper at step 2, repeat the experiment for 20 times. |
| Execution Constraints | Any grasping strategy can be utilized while grasping the objects. The grasping strategy can differ from object to object. However, the grasping strategy and the motion applied for SP1 should be kept the same for SP2, SP3 and SP4 for each object. It is not allowed to move the object to the edges of the planar object or the table. |

# Gripper Assessment Benchmark

| | |
|---|---|
| Reference No / Version | B-GA-0.01 |
| Authors | Berk Calli, Aaron Dollar |
| Institution | Yale University |
| Contact information | berk.calli@yale.edu |
| Adopted Protocol | Gripper Assessment Protocol (P-GA-0.01) |
| Scoring | Fill the attached table with the following rules:<br>For each set points, and for each object category other than articulated objects:<br>• After step 5 of the procedure, if the object is not dropped and if no visible motion of the object is detected within the gripper during steps 4 and 5, the gripper gets 2 points. If the object is not dropped, but visible object motion is detected during steps 4 and 5, the gripper gets 1 point.<br>• If the object is still not dropped after step 7 and no visible motion of the object is detected within the grip during steps 6 and 7, add 2 more points. If the object is not dropped, but visible object motion is detected during steps 6 and 7 add 1 point. Write the score to the corresponding cell.<br>For articulated objects:<br>• If no part of the object is touching the table after step 5, the grasp is considered successful. Add 0.5 points for each successful grasp. |
| To submit | • Scoring table (attached to the benchmark)<br>• Detailed descriptions of the gripper (or a reference that provide detailed descriptions).<br>• Detailed comments on:<br>o The advantages of the gripper.<br>o The disadvantages of the gripper.<br>o Reasons for the failed grasps. |

# Scoring Table for Gripper Assessment

| | |
|---|---|
| Gripper name/model: | |
| Manufacturer: | |
| Name/Institution: | |
| Contact info: | |

| | | SP1 | SP2 (x offset) | SP3 (y offset) | SP4 (z offset) |
|---|---|---|---|---|---|
| Round Objects | Soccer Ball | | | | |
| | Softball | | | | |
| | Tennis ball | | | | |
| | Racquetball | | | | |
| | Golf ball | | | | |
| | Marble XL | | | | |
| | Marble L | | | | |
| | Marble M | | | | |
| | Marble S | | | | |
| Flat Objects | Washer 1 (largest) | | | | |
| | Washer 2 | | | | |
| | Washer 3 | | | | |
| | Washer 4 | | | | |
| | Washer 5 | | | | |
| | Washer 6 | | | | |
| | Washer 7 (smallest) | | | | |
| | Credit card | | | | |
| Tools | Pen | | | | |
| | Scissors | | | | |
| | Hammer | | | | |
| | Screwdriver | | | | |
| | Driller | | | | |
| | Peg XL | | | | |
| | Peg L | | | | |
| | Peg M | | | | |
| | Peg S | | | | |
| Articulated | Chain | | | | |
| | Rope | | | | |

# Protocol for Table Setting

| | |
|---|---|
| Reference No / Version | P-TS-0.01 |
| Authors | Berk Calli |
| Institutions | Yale University |
| Contact information | berk.calli@yal.edu |
| Purpose | To assess the pick and place ability of a robotic system. |
| Task Description | Set up a table by placing kitchen inventories to predefined places. |
| Setup Description | **<u>List of objects and their descriptions:</u>**<br>The following objects from the YCB object set are used: The mug, the plate, the fork, the spoon, the knife, the bowl and the white table cloth. |
| | **<u>Initial poses of the objects</u>**<br>The initial positions of each object are defined with a printable layout provided in http://rll.eecs.berkeley.edu/ycb/. Print the layout to an A3 paper, and place it on the table. |
| | **<u>Description of the manipulation environment:</u>**<br>Conduct the experiment on the white table cloth provided by the object set. Avoid background clutter. The target positions of the objects are defined with a printable layout provided in http://rll.eecs.berkeley.edu/ycb/. The final position of each object is designated with a different colored region. Print the layout to an A3 paper, and place it next to the first printable layout as explained on the website. |
| Robot/Hardware Description | **<u>Targeted robots/hardware:</u>**<br>The protocol is designed for robots that have on board sensors. No sensors fixed to the environment should be used. |
| | **<u>Initial state of the robot with respect to the setup:</u>**<br>The manipulator of the robot can start from any initial configuration that is at least 10 cm away from the target objects. |
| | **<u>Prior information provided to the robot:</u>**<br>In order to avoid the use of object recognition algorithms and to evaluate only the manipulation performance, the object-color order is known to the robot: From left to right, the first object should be placed on the yellow region, the second object to the magenta region and the third object to the green region, forth object to the cyan region, fifth object to the blue region and sixed object to the red region.<br>If a grasp synthesis algorithm for known objects is used, the models provided by the YCB model and object set can be utilized. If an algorithm for unknown objects is used, no *a priori* shape information should be provided to the robot. |
| Procedure | 1. Place the objects to the initial positions indicated by the printable layout.<br>2. Run the system. |
| Execution Constraints | None. |

# Benchmark for Table Setting

| | |
|---|---|
| Reference No / Version | B-TS-0.01 |
| Authors | Berk Calli |
| Institution | Yale University |
| Contact information | berk.calli@yale.edu |
| Adopted Protocol | Protocol for Table Setting (P-TS-0.01) |
| Scoring | • For each object that is placed fully inside the correct colored area, the system gets 4 points.<br>• For each object that touches the correct color, but has some part outside the colored area, the system gets 2 points.<br>• For each object that is grasped and lifted successfully, but placed totally outside the correct color area, the system gets 1 point.<br>• Sum up the points for the overall score. |
| To Submit | • The resulting score.<br>• System Description<br>• Provide detailed comments on:<br>o What makes the system successful?<br>o What makes the system fail?<br>o What kind of improvements are necessary in<br>▪ the hardware design,<br>▪ the grasp synthesis algorithm,<br>▪ the manipulation strategy,<br>▪ and semantic modeling of the task? |

**APPENDIX B.4 BLOCK PICK AND PLACE PROTOCOL AND BENCHMARK:**

# Block Pick and Place Protocol

| | |
|---|---|
| Reference No / Version | P-BP-0.01 |
| Authors | Aaron Walsman, Siddhartha Srinivasa |
| Institution | CMU |
| Contact information | aaronwalsman@gmail.com |
| Purpose | Assess the dexterity of a robotic manipulator using a pick-and-place task. |
| Task Description | Arrange blocks into a specified pattern. |
| Setup Description | **<u>Description of the manipulation environment:</u>**<br>Print the template provided at http://rll.eecs.berkeley.edu/ycb/, and place it in front of the robot. The template signifies the target final positions of the blocks. |
| | **<u>List of objects and their descriptions:</u>**<br>Eight wooden blocks that are provided by YCB object and models set (as a part of the Box and Blocks Test) are used. |
| | **<u>Initial poses of the objects:</u>**<br>The blocks are placed on the table randomly but at least ten centimeters away from the printed template. |
| Robot/Hardware/Subject Description | **<u>Targeted robots/hardware/subjects:</u>**<br>Any robotic manipulator. |
| | **<u>Initial state of the robot/hardware/subject with respect to the setup:</u>**<br>The robot may start in any configuration such that the manipulator is at least ten centimeters from the blocks and template. |
| | **<u>Prior information provided to the robot/hardware/subject:</u>**<br>The starting configuration of the template and blocks can either be specified a-priori, or perceived using on-board sensors. |
| Procedure | 1) Place the blocks as described.<br>2) Run the system. |
| Execution Constraints | None. |

# Block Pick and Place Benchmark

| | |
|---|---|
| Reference No / Version | B-BP-0.01 |
| Authors | Aaron Walsman |
| Institution | CMU |
| Contact information | aaronwalsman@gmail.com |
| Adopted Protocol | Block Pick and Place Protocol (P-BP-0.01) |
| Scoring | The robot is awarded one point for each block that is touching a unique target area on the printed template. Two blocks touching the same target area does not score two points. Additionally, if the robot is awarded one additional point for any block that is entirely contained within the target area (does not touch the border of the target area). Finally, the robot loses one point for any block that is either partially or fully off of the printed template. |
| To submit | • Score.<br>• Provide detailed comments on:<br>○ What makes the system successful?<br>○ What makes the system fail?<br>○ What kind of improvements are necessary? |

APPENDIX B.5 PEG INSERTION LEARNING ASSESSMENT PROTOCOL AND BENCHMARK:

# Peg Insertion Learning Assessment Protocol

| | |
|---|---|
| Reference No / Version | P-PI-0.01 |
| Authors | Arjun Singh, Pieter Abbeel |
| Institution | UC Berkeley |
| Contact information | arjun810@gmail.com, pabbeel@cs.berkeley.edu |
| Purpose | To assess the learning performance of a robot by the task of inserting a single peg to a hole. |
| Task Description | Insert a peg to a hole with a learned policy. |
| Setup Description | **<u>Description of the manipulation environment:</u>** |
| | **<u>List of objects and their descriptions:</u>**<br>9-peg hole test provided by YCB Object and model set is used. |
| | **<u>Initial poses of the objects:</u>**<br>The robot gripper holds the peg in a pose specified by the user. The pose of the peg with respect to the gripper is kept the same for all executions.<br>A *home position* for the peg board is defined by the user.<br>20 perturbed initial positions are defined for the peg board as follows:<br>• Two perturbation levels are defined as 5mm perturbation and 10mm perturbation.<br>• The peg board is moved away from the home position following a random direction with the specified amount of perturbation. This process is repeated 10 times for each perturbation level, and the positions are recorded. |
| Robot/Hardware/Subject Description | **<u>Targeted robots/hardware/subjects:</u>**<br>Any robotic manipulator that can translate the peg in x, y and z directions in the Cartesian space. |
| | **<u>Initial state of the robot/hardware/subject with respect to the setup:</u>**<br>The initial pose of the manipulator is set in such a way that the initial position of the peg is 10 cm above the target hole. |
| | **<u>Prior information provided to the robot/hardware/subject:</u>**<br>The home position of the peg board (thus the hole) is known by the robot. The perturbation amount and direction are unknown to the robot. |
| Procedure | Four execution durations are defined as 0.5 seconds, 1 second, 3 second and 5 second.<br>The procedure is as follows:<br>1) Place the peg board to the initial position.<br>2) Move the robot to the initial position.<br>3) Run the peg insertion procedure.<br>4) Terminate the procedure after the execution duration.<br>• Execute this procedure 10 times for all four execution durations by setting the initial position of the peg board to the home position (40 experiments).<br>• Execute this procedure once for each (recorded) perturbed initial positions for all four execution durations (80 experiments in total). |
| Execution Constraints | None. |

# Peg Insertion Learning Assessment Benchmark

| | |
|---|---|
| Reference No / Version | B-PI-0.01 |
| Authors | Arjun Singh, Pieter Abbeel |
| Institution | UC Berkeley |
| Contact information | arjun810@gmail.com, pabbeel@cs.berkeley.edu |
| Adopted Protocol | Peg Insertion Learning Assessment Protocol (P-PI-0.01) |
| Scoring | Fill the attached scoring table with the number of successful trials. |
| To Submit | • Scoring table.<br>• System Description.<br>• Provide detailed comments on:<br>○ Utilized learning technique.<br>○ What makes the system successful?<br>○ What makes the system fail?<br>○ What kind of improvements are necessary? |

# Scoring table for Peg Insertion Learning Assessment Benchmark

| | Execution time | | | |
|---|---|---|---|---|
| | 0.5 sec | 1 sec | 3 secs | 5 secs |
| No Perturbation | /10 | /10 | /10 | /10 |
| 5 mm Perturbation | /10 | /10 | /10 | /10 |
| 10 mm Perturbation | /10 | /10 | /10 | /10 |

## Appendix C. Benchmarking Results:

### Appendix C.1 Pitcher-Mug Benchmarking Results

| Score | Trial | Pitcher Before | Cup Before | Pitcher After | Cup After | Score |
|---|---|---|---|---|---|---|
| | 1 | .475 | .115 | .280 | .305 | 0.974 |
| | 2 | .470 | .115 | .265 | .320 | 1.0 |
| | 3 | .475 | .115 | .250 | .330 | 0 |
| | 4 | .470 | .115 | .275 | .305 | 0.974 |
| | 5 | .470 | .115 | .315 | .270 | 0 |
| | 6 | .520 | .115 | .250 | .385 | 1.0 |
| | 7 | .525 | .115 | .330 | .305 | 0.974 |
| | 8 | .520 | .115 | .315 | .320 | 1.0 |
| | 9 | .525 | .115 | .265 | .370 | .981 |
| | 10 | .520 | .115 | .275 | .355 | 0.980 |
| | Avg | .497 | .115 | .282 | .3265 | 0.7884 |
| System Description | Robot platform: HERB (*Siddhartha Srinivasa, Dmitry Berenson, Maya Cakmak, Alvaro Collet Romea, Mehmet Dogar, Anca Dragan, Ross Alan Knepper, Tim D. Niemueller, Kyle Strabala, J Michael Vandeweghe, and Julius Ziegler, "HERB 2.0: Lessons Learned from Developing a Mobile Manipulator for the Home," Proceedings of the IEEE, Vol. 100, No. 8, July, 2012, pp. 1-19)*<br>Gripper: Barrett BH280 (http://web.barrett.com/support/BarrettHand_Documentation/BH8-280_Datasheet.pdf) | | | | | |
| Comments | The table above shows the results of the Pitcher-Mug benchmark on the HERB robot platform using the CBiRRT planner with TSR constraints. In this test, the locations of the pitcher and mug were known a priori. Due to the robot's large hands we were unable to make use of the pitcher's handle and instead opted to pour by grasping the bottom of the pitcher cylinder itself. For each test run the robot planned to an arm configuration offset slightly from the pitcher, then ran an open loop action to push the hand towards the pitcher and close the hand. Once grasped, the robot lifted the pitcher and planned a pouring motion using the CBiRRT planner with TSR constraints to rotate the pitcher about a fixed point in front of the spout. We did not have the ability to estimate the pitcher mass in the hand to determine pouring progress, so we instead waited a fixed number of seconds before returning the pitcher to an upright position and placing it back on the table. On one run the robot was not able to replace the pitcher in an upright position. In another, it did not manage to fill the cup halfway. Both runs scored zero. Because we used a randomized planner for most of the motions, the path taken between endpoints is nondeterministic. This is one potential source of failure as occasionally the planner will find one path to an intermediate goal position, but then be unable to find a path to the next goal. This is indeed what happened in failure case on trial 3 above. The planner found a trajectory to get into the pour position, but was then unable to find a trajectory to set the pitcher back down. The other failure case in which the mug was not filled was caused by the pitcher not being tilted far enough when pouring the water. This was also a problem with planning. When setting up this experiment, we built in a backup mechanism whereby if the robot was unable to find a plan to tilt the pitcher all the way, it would attempt to plan to a different pose that was almost as far. This increased reliability, but also led to this failure case in which the water was not poured out all the way. The HERB robot performed well on this protocol in the non-failure cases. The results show that the robot's greatest room for improvement is in reliability rather than precision. | | | | | |

### APPENDIX C.2 GRIPPER BENCHMARKING RESULTS

The YCB Gripper Benchmark is applied to two grippers designed in Yale GRAB lab: Gripper Model T, Gripper Model T42. The results are presented below.

## Scoring Table for Gripper Assessment

| | |
|---|---|
| Gripper name/model: | Model T<br>*(Raymond R. Ma, Lael U. Odhner, Aaron M. Dollar "A Modular, Open-Source 3D Printed Underactuated Hands," Proceedings of the 2013 IEEE International Conference on Robotics and Automation (ICRA), Karlsruhe, Germany, May 6-10, 2013.)* |
| Manufacturer: | Yale GRAB LAB |
| Name/Institution: | Yale GRAB LAB |
| Contact info: | berk.calli@yale.edu |

| | | SP1 | SP2 (x offset) | SP3 (y offset) | SP4 (z offset) |
|---|---|---|---|---|---|
| Round Objects | Soccer Ball | 0 | 0 | 0 | 0 |
| | Softball | 4 | 3 | 2 | 2 |
| | Tennis ball | 4 | 4 | 4 | 4 |
| | Racquetball | 4 | 4 | 4 | 4 |
| | Golf ball | 0 | 0 | 0 | 0 |
| | Marble XL | 0 | 0 | 0 | 0 |
| | Marble L | 0 | 0 | 0 | 0 |
| | Marble M | 0 | 0 | 0 | 0 |
| | Marble S | 0 | 0 | 0 | 0 |
| Flat Objects | Washer 1 (largest) | 0 | 0 | 0 | |
| | Washer 2 | 0 | 0 | 0 | |
| | Washer 3 | 0 | 0 | 0 | |
| | Washer 4 | 0 | 0 | 0 | |
| | Washer 5 | 0 | 0 | 0 | |
| | Washer 6 | 0 | 0 | 0 | |
| | Washer 7 (smallest) | 0 | 0 | 0 | |
| | Credit card | 4 | 4 | 4 | |
| Tools | Pen | 0 | 0 | 0 | 0 |
| | Scissors | 0 | 0 | 0 | 0 |
| | Hammer | 0 | 0 | 0 | 0 |
| | Screwdriver | 0 | 0 | 0 | 0 |
| | Driller | 0 | 0 | 0 | 0 |
| | Peg XL | 4 | 4 | 4 | 4 |
| | Peg L | 4 | 4 | 4 | 4 |
| | Peg M | 4 | 4 | 4 | 4 |
| | Peg S | 2 | 0 | 0 | 0 |
| Articulated | Chain | 7 | | | |
| | Rope | 10 | | | |

| | | |
|---|---|---|
| Score | Round objects | 43/187 |
| | Flat objects | 12/96 |
| | Tools | 50/144 |
| | Articulated | 17/20 |
| | Total | 122/447 |
| Comments | Comments: The Model T gripper provided stable grasps for a limited range of object sizes. It was unsuccessful for flat objects with round edges (washers), but maintained robust results for flat objects with straight edges. For most of the tool items, the gripper failed to provide a stable grasps. The reason of failure while grasping large objects appears to be due to the flexure joints at the finger bases: When the object gets bigger, these flexure joints apply torsional forces on the object which degrades the stability of the grasp. While grasping small objects, the ability of the gripper fingers to interlace works as a disadvantage: The objects smaller than the gaps between the fingers cannot be grasped with this gripper. The ability to interlace brings an advantage while grasping articulated objects: While grasping the chain and the rope, the gripper fingers can tightly wrap around the object and provide stable grasps. Interlacing appears to be an advantage while grasping objects like driller and hammer: The fingers can wrap around these objects and provide large amount of contact surfaces. However, the flexure joints at the finger bases are not able to provide enough stiffness to lift these heavy objects. While grasping the clippers, it is observed that the gripper can take significant advantage of the concavities of the object. To summarize, even though the flexure joints at the finger bases provide adaptability, they are not able to supply the stiffness to grasp large and heavy objects. The performance of this gripper can be improved significantly, if these base flexure joints are replaced with pin joints. | |

# Scoring Table for Gripper Assessment

| | |
|---|---|
| Gripper name/model: | Model T42<br>*(Raymond R. Ma, Lael U. Odhner, Aaron M. Dollar "A Modular, Open-Source 3D Printed Underactuated Hands," Proceedings of the 2013 IEEE International Conference on Robotics and Automation (ICRA), Karlsruhe, Germany, May 6-10, 2013.)* |
| Manufacturer: | Yale GRAB LAB |
| Name/Institution: | Yale GRAB LAB |
| Contact info: | berk.calli@yale.edu |

| | | SP1 | SP2 (x offset) | SP3 (y offset) | SP4 (z offset) |
|---|---|---|---|---|---|
| Round Objects | Soccer Ball | 3 | 2 | 3 | 3 |
| | Softball | 4 | 4 | 4 | 4 |
| | Tennis ball | 4 | 4 | 4 | 4 |
| | Racquetball | 4 | 4 | 4 | 4 |
| | Golf ball | 4 | 4 | 4 | 4 |
| | Marble XL | 4 | 4 | 4 | 4 |
| | Marble L | 4 | 4 | 4 | 4 |
| | Marble M | 4 | 4 | 4 | 4 |
| | Marble S | 4 | 4 | 4 | 4 |
| Flat Objects | Washer 1 (largest) | 4 | 4 | 4 | |
| | Washer 2 | 4 | 4 | 4 | |
| | Washer 3 | 4 | 4 | 4 | |
| | Washer 4 | 4 | 4 | 4 | |
| | Washer 5 | 4 | 4 | 4 | |
| | Washer 6 | 4 | 4 | 4 | |
| | Washer 7 (smallest) | 0 | 0 | 0 | |
| | Credit card | 4 | 4 | 4 | |
| Tools | Pen | 4 | 4 | 4 | 4 |
| | Scissors | 4 | 4 | 4 | 4 |
| | Hammer | 0 | 0 | 0 | 0 |
| | Screwdriver | 4 | 4 | 4 | 4 |
| | Driller | 3 | 3 | 3 | 3 |
| | Peg XL | 4 | 4 | 4 | 4 |
| | Peg L | 4 | 3 | 3 | 4 |
| | Peg M | 4 | 4 | 4 | 3 |
| | Peg S | 4 | 4 | 4 | 4 |
| Articulated | Chain | 2 | | | |
| | Rope | 10 | | | |

| | | |
|---|---|---|
| Score | Round objects | 182/187 |
| | Flat objects | 84/96 |
| | Tools | 91/114 |
| | Articulated | 12/20 |
| | Total | 379/447 |
| Comments | The Model T42 gripper provided stable grasps for a large range of object sizes and shapes. It is observed that it can provide power grasps by adapting to large shapes, and can provide precision grasps while grasping small objects thanks to its finger design. By utilizing its nails, the gripper was able to grasp very tiny and flat objects (i.e. small washers and credit card) robustly. The main disadvantage appears to be the non-interlacing fingers. For objects like hammer and driller, the non-interlacing fingers prevent to acquire firm grasps due to insufficient contact surfaces. The same disadvantage also applies while grasping articulated objects. It is also observed that concavities on the object surface sometimes leads to insufficient contact which prevents a firm grasp. This has been observed while grasping the clippers. | |

## Appendix C.3 Block Pick And Place Benchmarking Results:

| Trial | Blocks Touching Target Areas | Blocks Entirely Within Target Areas | Blocks Partially Off Printed Template | Score |
|---|---|---|---|---|
| 1 | 7 | 0 | 1 | 6 |
| 2 | 7 | 1 | 1 | 7 |
| 3 | 6 | 0 | 1 | 5 |
| 4 | 0 | 0 | 8 | -8 |
| 5 | 1 | 0 | 7 | -6 |
| 6 | 7 | 0 | 0 | 7 |
| Avg | 4.67 | 0.17 | 3.0 | 1.83 |
| System Description | Robot platform: HERB (*Siddhartha Srinivasa, Dmitry Berenson, Maya Cakmak, Alvaro Collet Romea, Mehmet Dogar, Anca Dragan, Ross Alan Knepper, Tim D. Niemueller, Kyle Strabala, J Michael Vandeweghe, and Julius Ziegler, "HERB 2.0: Lessons Learned from Developing a Mobile Manipulator for the Home," Proceedings of the IEEE, Vol. 100, No. 8, July, 2012, pp. 1-19)*<br>Gripper: Barrett BH280 (http://web.barrett.com/support/BarrettHand_Documentation/BH8-280_Datasheet.pdf) | | | |
| Comments | The table above shows six test runs of the block placement protocol on the HERB robot platform using the CBiRRT planner with TSR constraints. Because of the relatively large hands of the HERB robot and the small size of the blocks, we developed a block grasping strategy that consisted of using CBiRRT to plan the end effector to a position above and slightly behind a selected block. The system then runs a series of open-loop actions that lower the hand to the table height and push the end-effector forward while the hand is in a position designed to funnel the block into a graspable position. The hand then closes, grasping the block and the end-effector pulls away from the table again. In order to place the block, CBiRRT plans the end-effector to a position above the desired final location of the block, then another open-loop action lowers the hand to the table and gently sets the block down to avoid unnecessary bouncing. We found this strategy was largely successful at placing the blocks such that they at least touched the target locations, however due to uncertainties introduced by the open-loop motions, we were not able to reliably place the blocks such that they were entirely within the target squares. In these results, trial 4 and 5 encountered irrecoverable conditions early into the run, which dropped the overall average substantially. If greater reliability could be achieved, an average score above 6.0 would be reasonable. Note that this is well below the 16.0 perfect score that would awarded if all blocks we placed entirely within their respective target areas. In order to achieve this score, more development would be required to achieve a grasp and place action that resulted in less uncertainty and more repeatability. For these tests, the failure cases were caused by planning failures with our randomized planner in which no paths were found either to pick the block up or to place it. This is partially because we specified relatively small goal regions for these actions, but also because some of the target regions were close to the edge of the robot's reachable workspace. The results provide a reasonable assessment of the robot's precision and repeatability when performing this task, and revealed a clear path towards improvement. | | | |

### APPENDIX C.4 PEG INSERTION LEARNING BENCHMARKING RESULTS:

| | Execution time | | | |
|---|---|---|---|---|
| | 0.5 sec | 1 sec | 3 secs | 5 secs |
| No Perturbation | 7/10 | 9/10 | 10/10 | 10/10 |
| 5 mm Perturbation | 2/10 | 4/10 | 4/10 | 4/10 |
| 10 mm Perturbation | 0/10 | 0/10 | 0/10 | 0/10 |
| | | | | |
| System Description | PR2 (https://www.willowgarage.com/pages/pr2/overview) | | | |
| Comments | **Utilized learning technique:** In order to achieve the manipulation task, a linear-Gaussian controller is learned. The state consists of the joint angles and angular velocities of the robot, and the positions and velocities of three points in the space of the end effector (3 points in order to fully define a rigid body configuration).The objective during training depends on the distance between these points and the target position (measured before the random, unknown perturbations are added, which help to train a more robust controller). No information is available to the controller at test time except joint angles, angular velocities, and the positions & velocities of those three points. A random perturbation in the horizontal plane is sampled from a normal distribution with standard deviation of 3 mm for each trial. | | | |

### APPENDIX C.5 BOX AND BLOCKS TEST RESULTS:

**Utilized approach:** The robot picks a location from a uniform distribution over the box and attempts to pick up a block. The gripper's pose aligns with the length of the box. The gripper is then closed, and checked if it is fully closed. If the gripper closes fully, this means no blocks have been grasped and therefore the robot chooses a new location to attempt another pick. The robot repeats this heuristic until the gripper is not fully closed. When a grasp is detected, the robot moves to the destination box and releases the box. By using this heuristic, we run 10 experiments of 2 minutes each, and the results are presented in the table below.

| System Description | PR2 (https://www.willowgarage.com/pages/pr2/overview) | | | | | | | |
|---|---|---|---|---|---|---|---|---|
| Trial | Miss** | Pick 1* | Pick 2* | Pick 3* | Pick 4* | Successful Transfers | Blocks Transfered | Transfer fail*** |
| 1 | 6 | 4 | 4 | 0 | 0 | 8 | 12 | 0 |
| 2 | 4 | 3 | 6 | 0 | 0 | 9 | 15 | 1 |
| 3 | 5 | 6 | 2 | 0 | 0 | 8 | 10 | 1 |
| 4 | 7 | 3 | 2 | 0 | 1 | 6 | 11 | 1 |
| 5 | 7 | 5 | 2 | 0 | 0 | 7 | 9 | 1 |
| 6 | 6 | 7 | 1 | 0 | 0 | 8 | 9 | 0 |
| 7 | 7 | 4 | 3 | 0 | 0 | 7 | 10 | 1 |
| 8 | 4 | 9 | 0 | 0 | 0 | 9 | 9 | 0 |
| 9 | 12 | 0 | 4 | 0 | 0 | 4 | 8 | 1 |
| 10 | 7 | 0 | 6 | 1 | 0 | 7 | 15 | 1 |
| | | | | | | | | |
| Mean | 6.5 | 4.1 | 3 | 0.1 | 0.1 | 7.3 | 10.8 | 0.7 |
| Standard devation | 2.15 | 2.7 | 1.89 | 0.3 | 0.3 | 1.41 | 2.35 | 0.45 |
| *Pick x: the number of times the robot successfully picked and dropped off x blocks.<br>**Miss: the number of times the gripper closed and there was no block in it.<br>***Transfer fail: the number of times the robot fails to pick up a block and performs a drop off anyways. | | | | | | | | |